\documentclass[preprint,12pt,authoryear]{elsarticle}

\def\appendixname{}

\usepackage{amssymb}
\usepackage{amsmath}
\usepackage{bm}
\usepackage{hyperref}
\hypersetup{colorlinks = true,linkcolor = blue,anchorcolor =red,citecolor = blue,filecolor = red,urlcolor = red,
            pdfauthor=author}
\usepackage{bibunits}

\defaultbibliographystyle{elsarticle-harv} 
\defaultbibliography{references_manual}

\usepackage{geometry}
\usepackage{dirtytalk}

\usepackage{tabularx}
\usepackage{booktabs}
\usepackage{siunitx}

\usepackage{multirow}
\usepackage[table,dvipsnames]{xcolor}
\usepackage{etoolbox}
\robustify\bfseries
\robustify\itshape
\usepackage{longtable}

\usepackage{algorithm}
\usepackage{algpseudocode}
\algrenewcommand\algorithmicrequire{\textbf{Input:} }
\algrenewcommand\algorithmicreturn{\textbf{Output:} }

\usepackage{verbatim}

\makeatletter
\@tempswafalse
\def\@tempa{draft}
\@for\next:=\@classoptionslist\do
  {\ifx\next\@tempa\@tempswatrue\fi}
\if@tempswa %
    \renewenvironment{longtable}{\comment}{\endcomment}
    \fi
\makeatother

\usepackage{cleveref}
\Crefname{appendix}{Supplementary Information}{Supplementary Information}
\crefname{appendixfigure}{Supplementary Figure}{Supplementary Figures}
\crefname{appendixtable}{Supplementary Table}{Supplementary Tables}
\begin{document}

\sisetup{detect-all = true}

\begin{frontmatter}

\title{Generative clinical time series models trained on moderate amounts of patient data are privacy preserving} %

\author[ptb]{Rustam Zhumagambetov} %
\cortext[cor1]{Corresponding author}
\author[charite]{Niklas Giesa} %
\author[charite]{Sebastian D.~Boie} %
\author[tub,ptb,charite]{Stefan Haufe\corref{cor1}} %
\ead{haufe@tu-berlin.de}
\affiliation[ptb]{organization={Mathematical Modelling and Data Analysis Department, Physikalisch-Technische Bundesanstalt (PTB)},%
            addressline={Abbestraße~2-12}, 
            city={Berlin},
            postcode={10587}, 
            country={Germany}}

\affiliation[charite]{organization={Institute of Medical Informatics,
  Charité – Universitätsmedizin Berlin},%
            addressline={Invalidenstraße 90}, 
            city={Berlin},
            postcode={10117}, 
            country={Germany}}

\affiliation[tub]{organization={Uncertainty, Inverse Modeling and Machine Learning Group,
  Technische Universität Berlin},%
            addressline={Marchstr 23}, 
            city={Berlin},
            postcode={10587}, 
            country={Germany}}

\begin{abstract}
Sharing medical data for machine learning model training purposes is often impossible due to the risk of disclosing identifying information about individual patients. 
Synthetic data produced by generative artificial intelligence (genAI) models trained on real data is often seen as one possible solution to comply with privacy regulations. 
While powerful genAI models for heterogeneous hospital time series have recently been introduced, such modeling does not guarantee privacy protection, as the generated data may still reveal identifying information about individuals in the models' training cohort. Applying established privacy mechanisms to generative time series models, however, proves challenging as post-hoc data anonymization through k-anonymization or similar techniques is limited, while model-centered privacy mechanisms that implement differential privacy (DP) may lead to unstable training, compromising the utility of generated data.
Given these known limitations, privacy audits for generative time series models are currently indispensable regardless of the concrete privacy mechanisms applied to models and/or data.
In this work, we use a battery of established privacy attacks to audit state-of-the-art hospital time series models, trained on the public MIMIC-IV dataset, with respect to privacy preservation. Furthermore, the eICU dataset was used to mount a privacy attack against the synthetic data generator trained on the MIMIC-IV dataset. Results show that established privacy attacks are ineffective against generated multivariate clinical time series when synthetic data generators are trained on large enough training datasets. Furthermore, we discuss how the use of existing DP mechanisms for these synthetic data generators would not bring desired improvement in privacy, but only a decrease in utility for machine learning prediction tasks.
\end{abstract}


\end{frontmatter}

\begin{bibunit}
\section{Introduction}

Access to medical data is essential for research and innovation in medicine. It enables cross-hospital research cooperation~\citep{ross_importance_2012}, increases transparency in publications~\citep{sixto-costoya_emergency_2020}, and, in particular, facilitates the training of clinical prediction models using machine learning (ML). However, current privacy regulations such as GDPR \citep[General Data Protection Regulation,][]{GDPR2016a} or HIPAA \citep[Health Insurance Portability and Accountability Act,][]{hipaa_1996}  limit the dissemination of routine data acquired in hospitals. Even though these regulations typically allow the use of patient data 
for the greater societal good, providing access to possibly identifiable personal data faces great hurdles~\citep{edpb_2024}. This is especially the case for routinely collected data, for which no informed patient consent can be obtained, such as data acquired in intensive care units (ICU).

Anonymization and de-identification are possible solutions to enable routine data sharing ~\citep{gadotti_anonymization_2024}. However, anonymization cannot be fully automated and in some cases is not fully possible~\citep{zuo_data_2021}. De-identification is a process of removing direct personal identifiers, and is directly referenced in HIPAA~\citep{hipaa_1996}. Removing 18 identifiers allows the data to be used without HIPAA restrictions~\citep{45CFR164}. However, de-identification does not guarantee anonymity.

An alternative approach is the use of data synthesized by generative artificial intelligence (genAI) models trained on real patient data~\citep{lin_using_2020, theodorou_synthesize_2023}. Here, the idea is to substitute personal data with randomly generated data that preserve all relevant statistical aspects of real data in order to provide the utility of the original data for any desired purpose, while at the same avoiding any too close resemblance between individual generated samples and individual patient records contained in the training set.

While personal data are strictly regulated, synthetic data appear to provide a way to comply with privacy regulations while preserving utility for tasks such as machine learning model training. Moreover, genAI models can handle data types that are hard to anonymize without destroying utility, such as heterogeneous medical records ICUs~\citep{patharkar_predictive_2024,zhumagambetov_ghosts_2024}.

Modern ICUs continuously record patient data with high time resolution. These data often comprise vital signs,  laboratory results, medication records, and diagnostic information. When combined with historical hospital data, these records form comprehensive electronic health records (EHRs). 
ML models trained on EHRs have demonstrated notable success in predicting clinical outcomes such as acute kidney injury\citep{tomasev_clinically_2019-1}, perioperative complications~\citep{giesa_predicting_2024, giesa_applying_2024}, and patient mortality~\citep{lichtner_predicting_2021} ahead of time.
However, the lack of informed patient consent, and technical challenges in de-identifying or even anonymizing the data often prevent such data from being shared with researchers outside the medical center for machine learning model development~\citep{modra_informed_2014}. 

GenAI models, on the other hand, are not necessarily private by design. Depending on the training regime, distribution of training data, neural architecture, and selection of hyperparameters, generative models can exhibit a variety of failure modes that can lead to disclosure of private information not intended for public release~\citep{salimans_improved_2016}. This includes mode collapse~\citep{arnout_evaluation_2021, bhagyashree_study_2020}, where the model only captures a limited patient group/cohort compared to the original dataset, or dataset memorization~\citep{baiTrainingSampleMemorization2021}, where the model only outputs data in close vicinity to individual training samples. However, private information leakage can even occur even when there are no such apparent defects~\citep{hayesLOGANMembershipInference2018, yeom_privacy_2018}. 

Such failure modes of generative models also increase the risk that auxiliary information can used for mounting privacy attacks~\citep{sweeney_weaving_1997}. For example, the Washington State patient-level health dataset was partially re-identified by linking records to publicly available news reports\citep{sweeney_only_2015}. By matching demographic data and hospitalization details from news stories containing the word “hospitalized” with the state’s database, researchers uniquely and exactly identified  35 patients of the 81 news stories sampled. This demonstration of a linkage attack proved that even without names or addresses, the combination of ZIP codes, diagnoses, and admission dates was sufficient to put names to medical records, eventually forcing the state to restrict data access\citep{sweeney_only_2015}. Another privacy breach was identified in the audit of synthetic financial datasets, such as those analyzed by researchers involving GretelAI's tools and open-source models \citep{amanyashblog2024}. By identifying \say{PII Replay} -- where generative models inadvertently memorize and repeat verbatim strings from the training data -- researchers found that synthetic outputs contained real names, phone numbers, and email addresses found in public SEC filings. By matching these leaked identifiers with auxiliary public records, \citet{amanyashblog2024} demonstrated that supposedly "safe" synthetic data could explicitly unmask the private identities and financial associations of real-world individuals. Individual sources of information in both of these examples do not reveal private information on their own, but combining them with auxiliary non-private information provides private information to the attacker.

By decoupling synthetic data from real data, researchers can intuitively assume a lower risk of identity disclosure since no explicit link exists between the two. However, this assumption often overlooks subtle statistical cues that a generative model can leave in the synthetic data~\citep{hayesLOGANMembershipInference2018}, possibly allowing an attacker to infer an individual's membership in the training dataset. This type of privacy violation is known as a membership inference attack (MIA). MIAs are relevant in the medical context where disease-centric datasets are very common~\citep{heart_disease_45, diabetes_34, breast_cancer_wisconsin_(diagnostic)_17} and where membership disclosure may reveal sensitive private information with possible negative consequences for the individual (e.g., with respect to future employment).

To address these concerns, researchers have been advocating for the use of differential privacy (DP)~\citep{dworkPrivacyPreservingDataminingVertically2004}. Introduced by \citet{dworkPrivacyPreservingDataminingVertically2004}, DP is a mathematically rigorous definition of privacy. It defines a measure of \textit{disclosure}, which, intuitively, quantifies whether an adversary can differentiate between two databases that differ in a single element with log probability $\epsilon$. A number of algorithms have been proposed that satisfy this definition~\citep{mcsherry_mechanism_2007,blum_learning_2013,abadiDeepLearningDifferential2016}. DP provides robust theoretical privacy guarantees that do not depend on the context and size of the dataset. Specifically, if implemented correctly, DP protects individuals from membership inference and linkage attacks as described above~\citep{dwork_algorithmic_2014}. For instance, a dataset of smokers was analyzed, and smoking was linked to increased risk of heart disease. While DP does not prevent the disclosure of general findings about smoking, an individual’s risk of negative consequences through identification is no greater for participants in the dataset than for non-participants. DP does not prevent attackers from inferring certain general statistical associations between private and public attributes, though, since such associations are independent of the inclusion of an individual in the dataset.

First developed for preserving the privacy of database queries, the concept of DP has recently been adapted to discriminative~\citep{abadiDeepLearningDifferential2016} and generative machine learning algorithms~\citep{xie_differentially_2018}. The most frequent way to formulate DP in machine learning contexts is $(\epsilon, \delta)$-differential privacy~\citep{vaudenay_our_2006}, where $\epsilon$ denotes the privacy budget and $\delta$ indicates the probability of not achieving $\epsilon$ guarantees. 
An algorithm is $(\epsilon, \delta)$-differentially private if an examination of the algorithm's output cannot reliably tell, bounded by probability $\exp(\epsilon)$, whether any specific individual’s information was included in the dataset or not, allowing a failure probability of $\delta$.
An increase in the privacy budget corresponds to a decrease in the level of privacy protection afforded by the model. \citet{dwork2011a} suggests $\epsilon$ of \say{0.01, 0.1, or in some cases, ln 2 or ln 3}. \citet{lee_how_2011} provide methodology for choosing $\epsilon$.%

However, the incorporation of differential privacy into machine learning algorithms is not straightforward. Differentially private ML algorithms are extremely sensitive to the tuning of the privacy hyperparameters~\citep{kulynych2024attack}, and the practical use of these parameters presents difficulties in interpretation~\citep{hsu_differential_2014,lowy_why_2024}. Moreover, the common approach of implementing DP for ML models through modifications of the training objective has been demonstrated to be insufficient to achieve theoretical DP privacy bounds in practice \citep{papernot2017semisupervised}.

Due to this insufficiency of privacy-promoting algorithmic tools to practically achieve differentially private generative ML models, empirical privacy audits are necessary. To this end, a number of privacy attacks have been designed~\citep{hayesLOGANMembershipInference2018,hilprechtMonteCarloReconstruction2019,vanbreugelMembershipInferenceAttacks2023}. Most importantly, MIA \citep{vanbreugelMembershipInferenceAttacks2023} aim to determine if a certain individual has been included in the training data of an ML model.
These attacks have been used to empirically assess DP properties of ML models trained on either real or synthetic data~\citep{annamalai_what_2024, kong_dp-auditorium_2024}. But such analyses have so far been limited to the image ~\citep{hayesLOGANMembershipInference2018} and tabular data~\citep{hilprechtMonteCarloReconstruction2019} domains. In contrast to these domains, high-dimensional medical time series are highly heterogeneous in terms of sampling intervals and observed data types, are characterized by complex non-linear spatio-temporal dependency structures, and are affected by structured missingness~\citep[e.g.,][]{patharkar_predictive_2024}, necessitating the development of dedicated generative AI tools for synthesizing such data ~\citep{theodorou_synthesize_2023,zhumagambetov_ghosts_2024}. 
The susceptibility of synthetic medical time series generated by these models to membership inference attacks has not been analyzed to date.

A fundamental challenge persists across all generative architectures: the mechanisms used to ensure faithfulness and mitigate issues like mode collapse—such as various divergence metrics—are inherently insufficient to provide robust privacy guarantees. Distributional metrics prioritize faithfulness over privacy; they ignore the degree of overfitting that ultimately facilitates MI attacks~\citep{yeom_privacy_2018}. While some works incorporate privacy considerations, evaluations are often restricted to basic distance metrics~\citep{sdmetrics} using only training and generated datasets, e.g., measuring distance to the closest record between real and synthetic sets, which lack formal privacy definitions or rigorous testing against realistic attack models.

To the authors' knowledge, this study is the first to audit existing synthetic hospital time series generation techniques using a comprehensive collection of realistic privacy attacks. 
More specifically, our investigation aims to address the following research questions:

\begin{enumerate}
    \item Can SOTA MIA successfully disclose whether the data of individual patients was used to train generative time series models, using only published synthetic dataset? %
    \item Do generative time series models developed without DP mechanism still offer protection against MIA? If so, how does the training sample size influence privacy protection?
    \item Is it possible to use a critical care dataset from one hospital to mount a privacy attack on synthetic data generators trained on data from another critical care dataset? We evaluate this on the well-known datasets eICU~\citep{pollard_eicu_2018} and MIMIC-IV~\citep{johnson_mimic-iv_2023}. %
\end{enumerate}

To fill these research gaps, this study offers the following contributions:
\begin{enumerate}
    \item We collect privacy attacks that have been used to conduct privacy attacks against synthetic data, contextualizing their relevance in hospital time series settings.
    \item We develop a framework for quantifying the ability of arbitrary synthetic time series generation methods to provide privacy protection using privacy attacks and quantitative metrics, and we provide source code for reproducibility.
    \item Using this framework, we perform a thorough privacy audit of four state-of-the-art models for hospital time series generation. We evaluate the results on two public intensive care datasets, MIMIC-IV \citep{johnson_mimic-iv_2023} and eICU~\citep{pollard_eicu_2018}.
\end{enumerate}

\section{Methods}

We first specify the cohort extraction procedures performed on the two datasets. Then, we briefly discuss the principles behind generative modeling and describe four generative time series models used in the experiments: GHOSTS, HALO, KoVAE, and Diffusion-TS. In the following subsection, we introduce the concept of membership inference attacks and categorize existing MI attacks into different types. Finally, we define the performance metrics used to assess the success of the privacy attacks and a metric for quantifying overfitting in generative models.

\subsection{Real data}
\paragraph{MIMIC-IV}
MIMIC-IV~\citep{johnson_mimic-iv_2023} is a comprehensive database containing de-identified records of over 300,000 hospitalized patients~\citep{johnson_mimic-iv_2023}. To retrieve EHR data from its tables, we utilized extraction routines provided in the official MIMIC-IV companion repository~\citep{johnson2018mimic,alistair_johnson_2024_13374956}. Data preprocessing involved filtering out outliers that fell out of physiological limits as defined by~\citet{harutyunyan_multitask_2019}. We extracted a set of $|F| = 9$ ICU time series variables: $F = \{$SOFA score, diastolic blood pressure, systolic blood pressure, heart rate, respiratory rate, SpO2, temperature, sodium, and glucose$\}$. These variables were chosen based on their relevance to clinical decision-making in intensive care settings~\citep{ferreira_serial_2001, yee_clinical_2023}. Basic statistics can be found in~\Cref{tab:features}. Data preparation steps can be found in \Cref{sec:data-preparation}.
 
We split the data into disjoint training, validation, and test folds based on patient identifiers.
We extracted $N_{\text{train}} = 21,057$ and $N_{\text{val}} = 1,282$ ICU stays from 16,320 and 1,000 patients for the training and validation splits, respectively, as well as $N_{\text{holdout}} = 22,463$ stays from 17,321 patients for the holdout split. 

Each ICU stay comprises a $|F| \times |T|$ matrix $\mathbf{d}_{\text{time}} = (d_{f,t})$ of time series data, where $f \in F$ indexes the feature and  $t \in T$ is a time index. In addition, each ICU stay is associated with static patient attributes $\mathbf{d}_{\text{attr}} = (g, a, m, l, r)$, where $g \in \{F, M\}$ is biological sex, $a \in \mathbb{N}^+$ is the patient age at the time of ICU admission, $m$ is marital status (single, married, divorced, widowed, missing), $l$ is length of stay in the ICU, and $r$ is reported race. The MIMIC training and test datasets are denoted by $\mathcal{D}^{\text{MIMIC}}_{\text{train}} = \{(\mathbf{d}_\text{time}, \mathbf{d}_\text{attr})_n: 0 < n < N_{\text{train}} + N_{\text{val}} \}$ and $\mathcal{D}^{\text{MIMIC}}_{\text{holdout}} = \{(\mathbf{d}_\text{time}, \mathbf{d}_\text{attr})_n: 0 < n < N_{\text{test}} \}$, where $\mathcal{D}^{\text{MIMIC}}_{\text{train}} \cap \mathcal{D}^{\text{MIMIC}}_{\text{holdout}} = \emptyset$. 

\paragraph{eICU} {The eICU database contains extensive ICU data collected from multiple hospitals across the United States~\citep{pollard_eicu_2018}. We extracted data using the ricu R package~\citep{bennett2023ricu}, applying the same selection criteria used for MIMIC-IV, as described in \Cref{sec:data-preparation}. This yielded 8,348 ICU stays from 8,163 unique patients. 
Demographic attributes were not extracted from eICU due to missing data on race and marital status. The collection resulted in a total of $|\mathcal{D}^{\text{eICU}}| = 8348$ ICU stays.}

\subsection*{Ethics Statement}

For MIMIC-IV and eICU datasets, data privacy and sharing regulations apply which are further defined at the PhysioNet platform~\citep{johnson_mimic-iv_nodate}.

\subsection{Generative models for multivariate time series data}

The purpose of generative models is to learn the probability distribution of a set of variables from training data in order to produce samples similar to real examples. Often, it is achieved by mapping a sample from a latent space with known simple distribution into a sample following the distribution of the training data. %
Suppose we have training data $\mathcal{D}_{\text{train}} = \{\mathbf{d}_i: \mathbf{d}_i \in \mathbb{D} \subseteq \mathbb{R}^k, 0 < i < N_{\text{train}}, k \in \mathbb{N}^+\}$ sampled from a distribution $P_R$.
To empirically estimate $P_R$, we define a parametric family of probability densities, $(P_{\bm{\Theta}})_{{\bm{\Theta}} \in \mathbb{R}^P}$, where $P \in \mathbb{N}^+$, \citep{arjovsky_wasserstein_2017}. The task is then to find the parameter vector $\bm{\Theta}$ that maximizes the log-likelihood of the training data, that is,
\begin{equation*}
\hat{\bm{\Theta}} = \text{arg} \; \text{max}_{\bm{\Theta}}\frac{1}{N_{\text{train}}}\sum_{i=1}^{N_{\text{train}}} \log P_{\bm{\Theta}} (\mathbf{d}_i).
\end{equation*}
Generative ML models implicitly represent $P_{\bm{\Theta}}$ by a learnable generator function $G_{\bm{\Theta}} : \mathcal{Z} \rightarrow \mathbb{D}$ that maps random inputs $\mathbf{z} \in \mathcal{Z} \subseteq \mathbb{R}^Q$ to samples $\Tilde{\mathbf{d}}$.

In time-series generation, the goal is to produce synthetic sequences ($d_{time}$) that accurately reflect the complex spatio-temporal dependencies—the relationships both between different variables and within individual variables over time—found in real data.

\paragraph*{Generator of Hospital Time Series (GHOSTS)}
GHOSTS~\citep{zhumagambetov_ghosts_2024} is a generative model based on the generative adversarial model (GAN) architecture. It augments the classical GAN loss function with structured sparsity penalties, thereby promoting GHOSTS's ability to model unevenly-sampled heterogeneous medical time series. A variant of GHOSTS with additional postprocessing to further increase the realism of the generated time series is GHOSTS-POST, also described in~\citet{zhumagambetov_ghosts_2024}.

\paragraph*{Hierarchical Autoregressive Language Model (HALO)}
HALO~\citep{theodorou_synthesize_2023} is a language model proposed for the generation of longitudinal, high-dimen\-sional EHR time series data in a hierarchical fashion. It combines a transformer model with an autoregressive masked linear model and is trained on tokens of discrete and discretized continuous values. Tokens are combined in visits that reflect diagnoses, procedures, and medications. The model is trained in an autoregressive fashion.

\paragraph*{Koopman Variational Autoencoder (KOVAE)}
KOVAE~\citep{naiman2024generative} is a generative framework that uses Variational Autoencoders (VAE) with a novel design of the model prior inspired by Koopman theory, representing latent conditional prior dynamics using a linear map. The use of a linear map as a prior allows the usage of spectral tools to impose constraints on the eigenvalues of the linear map, thus incorporating domain knowledge. KOVAE is particularly suited for irregularly sampled time series because it treats the latent dynamics as a continuous-operator problem rather than a discrete-step problem.

\paragraph*{Diffusion-TS}
Diffusion-TS~\citep{yuan2024diffusionts} is a diffusion-based framework for the generation of high-dimensional multivariate time series. It uses an encoder-decoder scheme with a disentangled temporal representation, decomposing the time series into trend and seasonal components. It helps to preserve the wave structure of medical time series. It also differs from vanilla diffusion models in that the model does not reconstruct the noise, but the sample directly in each diffusion step, which increases the faithfulness of the generated data.

\subsection{Membership inference attacks}
\label{sec:mia-eqs}

Synthetic data present a promising direction to enable the sharing of personal health data for statistical and machine learning analyses while preserving the anonymity of individuals who contributed their data. However, if a generative model is trained improperly leading, for instance, to overfitting, synthetic samples provide an adversary with a tool to detect if a certain person's data were used for model training. Such attacks are called membership inference attacks. Formally, the attacker computes $P(m=1 | x_i, \Theta)$, the probability that a record $x_i$ has been used to train ($m=1$) model $\Theta$. 
While initially developed for testing discriminative models, MIAs can also target generative models. In general, MIAs can be divided into two categories: density-based attacks and model-based attacks. Instances of the first category exploit potential overfitting of the generative model by detecting samples from the synthetic data that are overrepresented in certain regions of the generated data space compared to the density of training samples in the same region. Instances of the second category exploit knowledge about the model's density estimation process. In addition, general privacy attacks can be divided into white-box, partial black-box, and black-box approaches, depending on the attacker's knowledge. While in black-box attacks, only generated data are available to the attacker, partial black-box attacks also have access to an auxiliary/reference dataset that may or may not overlap with the training data. White-box attacks can additionally access the model parameters. White-box attacks are excluded from our current analysis as they assume a level of adversary knowledge that is seldom encountered in practical scenarios and can be easily avoided. On the other hand, auxiliary datasets can be obtained from public records, social media, or previous data breaches~\citep{desai_background_2022}.

\subsubsection{Density-based MIA}
Monte Carlo attacks~\citep{hilprechtMonteCarloReconstruction2019} (MCAttack) are black-box attacks that aims to exploit the overfitting behavior of generative models by detecting synthetic samples in the vicinity of the training data. This approach assumes that the probability of a record $x_i$ being used for training of model $G_\Theta$ is proportional to the probability that $x_i$ is surrounded by synthetic records $g \sim P_G$ generated by $G$,
\begin{equation}
  \label{eq:MCAttack_assumption}  P(m=1 | x_i, \Theta) \propto P(g \in U_\theta(x_i)),
\end{equation}
where $U_\theta(x_i)$ is a $\theta$-neighbourhood of a record $x_i$ defined as $U_\theta(x) = \left\{x^\prime | d(x, x^\prime) \leq \theta \right\}$, where $d$ is a distance.~\citet{hilprechtMonteCarloReconstruction2019} propose to use the median of the minimum distance to each record $x_i$ for all generated samples $g$ to determine $\theta$. The probability $P(g \in U_\theta(x))$ of a record $x$ being surrounded by synthetic records increases as more synthetic records $g$ are located near it:
\begin{equation}
    \label{eq:density-eq}
    P(g \in U_\theta(x)) = \mathbb{E}_{g\ \sim P_G} \left(\mathbf{1}_{g\in U_\theta(x)} \right) \;,
\end{equation}
and is approximated via the Monte Carlo method:
\begin{equation}
    \label{eq:mc-attack}
    \mathcal{A}_{MC-\theta} (x) = \frac{1}{N} \sum^N_{n=1} \mathbf{1}_{g_n\in U_\theta(x)}  \;,
\end{equation}
where $g_n$ is the n\textsuperscript{th} generated record and $\mathbf{1}_{g_n\in U_\theta(x)}$ is an indicator function, equal to 1 when $g_n\in U_\theta(x)$, and 0 otherwise. As per assumption \eqref{eq:MCAttack_assumption}, training records will have higher values of $\mathcal{A}_{MC-\theta} (x)$ than records not used for training $G$.

In contrast, \citet{chen_gan-leaks_2020} approximate~\Cref{eq:density-eq} using Parzen window density estimation as
\begin{equation}
    P(m=1 | x_i, \Theta) \approx \frac{1}{N} \sum^N_{n=1} \exp(-L (x, g_n)),
\end{equation}
where $L$ is a distance, 
and $N$ is the number of synthetic samples used. Practically, \citet{chen_gan-leaks_2020} approximate the probability that record $x$ is in the training dataset by the minimal distance across the five nearest synthetic data points:
\begin{equation}
    \label{eq:gan-leak}
    \mathcal{A}_{\text{GAN-Leak-Chen}} (x) = \underset{g_n \sim P_G}{\operatorname{argmin}_{n=1}^5}\  L(x, g_n).
\end{equation}

For the present study, we use instantiations of this attack proposed by \citet{chen_gan-leaks_2020} and \citet{vanbreugelMembershipInferenceAttacks2023}, referred to as $\mathcal{A}_{\text{GAN-Leak-Chen}}$ and $\mathcal{A}_{\text{GAN-Leak-Breugel}}$ (see \eqref{eq:gan-leak-breugel}), respectively, where
\begin{equation}
    \label{eq:gan-leak-breugel}
    \mathcal{A}_{\text{GAN-Leak-Breugel}} (x) = \operatorname{exp}(- \underset{g_n \sim P_G}{\operatorname{argmin}}\  L(x, g_n)) \;.
\end{equation}
In addition, \citet{chenGANLeaksTaxonomyMembership2020} proposed a calibrated version of the attack for the case that a dataset drawn from the same distribution as the training dataset, but with nonoverlapping elements, referred to as $D_{\text{ref}}$, is available: 
\begin{align}
    \label{eq:gan-leak-cal}
    \mathcal{A}_{\text{GAN-Leak-cal}} (x) &=\operatorname{sigmoid}(-L_{cal} (x, g_n)) \text{, where } \\
    L_{cal} (x, g_n) &=  L (x, g_n) - L (x, g^{\text{ref}}_n)
\end{align}
where 
$g^{\text{ref}}_n$ is the n\textsuperscript{th} record generated by some generative model $G^{\text{ref}}$ trained on $D_{\text{ref}}$. We use a variant of this attack proposed  by \citet{vanbreugelMembershipInferenceAttacks2023}, where samples $d_{\text{ref}} \in D_{\text{ref}}$ are used directly instead of synthetic samples $g^{\text{ref}}_n$.

In \citet{vanbreugelMembershipInferenceAttacks2023}, two further approaches, both referred to as DOMIAS, are proposed. The first approach uses only the generated data. Formally,
\begin{equation}
    \label{eq:domias-eq1}
     \mathcal{A}_{\text{DOMIAS-eq1}} (x) = f(P_G (x)) \;,
\end{equation}
where $P_G$ indicates the generator's output distribution and $f : \mathbb{R} \rightarrow [0,1]$ is a monotonically increasing scoring function. \citet{vanbreugelMembershipInferenceAttacks2023} argue that ~\Cref{eq:domias-eq1} is insufficient as an MIA since 1) it does not capture the intrinsic distribution of the real data, and 2) it is not invariant to bijective transformations of the domain. Based on this reasoning, they propose to weight ~\Cref{eq:domias-eq1} by the real data distribution:
\begin{equation}
    \label{eq:domias-eq2}
    \mathcal{A}_{\text{DOMIAS-eq2}} (x)  = f\left(\frac{P_G (x)}{P_R(x)}\right) \;,
\end{equation}
where $P_R$ is the true data distribution approximated from an auxiliary dataset. \citet{vanbreugelMembershipInferenceAttacks2023} use two versions of the attack that differ in the density estimation method, where Gaussian kernel density estimation (KDE) ~\citep{parzen_estimation_1962} with Scott's rule~\citep{scott_multivariate_1992} ($\mathcal{A}_\text{DOMIAS-KDE}$) and flow-based density estimation BNAF~\citep{bnaf19} ($\mathcal{A}_\text{DOMIAS-BNAF}$) are used to estimate $P_G$ and $P_R$.

\subsubsection{Model-based MIA}

\citet{hayesLOGANMembershipInference2018} design a MI attack against GAN models by exploiting the neural architecture of these models. Specifically, they propose several attacks that take advantage of the GAN's discriminator. The first attack requires access to the discriminator of the trained GAN, $D$:
\begin{equation}
    \label{eq:logan-wb}
    \mathcal{A}_{\text{Logan-wb}} (x) = D(x) \;,
\end{equation}
where higher value of $D(x)$ mean higher probability of $x$ being a part of training dataset.
The idea behind this attack is that, if the GAN is overfit, the discriminator will place a higher confidence on samples that were a part of the training set. Since this attack requires access to the trained discriminator, which can be easily omitted for publishing, we consider this attack unrealistic and applicable only to GANs and do not include it in our assessment.

The second attack requires access to an auxiliary dataset drawn from the same distribution as the training data. For example, in a medical setting, it could be a separate dataset gathered in the same hospital, or an openly available subset of the private dataset~\citep{johnson_mimic-iv_nodate}. The idea of this type of attack is to use a classification network $D_{\text{aux}}$, which is trained to discriminate between synthetic and auxiliary records, where the classifier output is then used as a score to conduct the attack:
\begin{equation}
    \label{eq:logan-pb}
    \mathcal{A}_{\text{Logan-pb}} (x) = D_\text{aux}(x) \;,
\end{equation}
where higher value of $D_\text{aux}(x)$ mean higher probability of $x$ being a part of training dataset.

\subsection{Privacy assessment}
\label{sec:privacy-assessment}

GHOSTS, HALO, KOVAE, and Diffusion-TS were trained on MIMIC-IV data $\mathcal{D}^{\text{MIMIC}}_{\text{train}}$ to generate synthetic data $\mathcal{D}_{\text{synth}}$ with $N_\text{synth}=20,000$. %
We applied postprocessing routines described in \citep{zhumagambetov_ghosts_2024} to the output of GHOSTS to produce the GHOSTS-POST synthetic dataset variant. We used the hyperparameters provided by the authors in the original implementations for all methods. Then, we used the resulting $\mathcal{D}_{\text{synth}}$ as well as $\mathcal{D}^{\text{MIMIC}}_{\text{train}}$ and $\mathcal{D}^{\text{MIMIC}}_{\text{holdout}}$ to conduct all privacy attacks described in previous section.

For black-box privacy attacks described in \Cref{sec:mia-eqs} only synthetic data $\mathcal{D}_{\text{synth}}$ were used. In contrast, for partial black-box attacks, both $\mathcal{D}_{\text{synth}}$ and part of $\mathcal{D}^{\text{MIMIC}}_{\text{holdout}}$ were used for mounting an attack. 
Furthermore, to address research question 2, $\mathcal{D}^{\text{eICU}}_{\text{holdout}}$ was separately used for mounting partial black-box attacks.

To mount the membership inference attacks, the following procedure~\citep{hayesLOGANMembershipInference2018, vanbreugelMembershipInferenceAttacks2023, chen_gan-leaks_2020} was used, which is summarized in \Cref{alg:workflow}. First, depending on the attack, auxiliary information $\mathcal{D}_{\text{aux}}$, that the attacker might have, was sampled from $\mathcal{D}^{\text{MIMIC}}_{\text{holdout}}$. Then, the test dataset $\mathcal{D}_{\text{test}}$, which was used for testing of the privacy attacks, was created by sampling $N_\text{train}$ records from the rest of $\mathcal{D}^{\text{MIMIC}}_{\text{holdout}}$ and combining them with $\mathcal{D}^{\text{MIMIC}}_{\text{train}}$. Afterwards, a membership inference attack was trained on synthetic data generated by each particular method's $\mathcal{D}_{\text{synth}}$. If the attack was of the partial black-box type, then $\mathcal{D}_{\text{aux}}$ was also used to train the attack. Then, the attack's scores were computed on the $\mathcal{D}_{\text{test}}$ and a threshold was picked as the median of all scores, according to established literature~\citep{vanbreugelMembershipInferenceAttacks2023,hayesLOGANMembershipInference2018,hilprechtMonteCarloReconstruction2019}. Here, the median shows the attacker's knowledge about the proportion of true samples in the testing dataset, which is unrealistic in practice but also represents a worst-case scenario for data publishers. An alternative could be choosing an arbitrary percentile, which shows a poor performance compared to the median method as demonstrated by \citet{hilprechtMonteCarloReconstruction2019}. The final steps include predicting labels and calculating the attacks' scores for the performance metrics.

Since both Diffusion-TS and KoVAE can only generate time series, membership inference attacks were conducted in two settings: time series only and static patient attributes only. In the time series setting, all five synthetic data generation models were compared. On the other hand, in the static patient attributes setting, HALO and GHOSTS were compared, since they can jointly generate both attributes and time series.

\begin{algorithm}
\caption{Membership inference attack}\label{alg:workflow}
\begin{algorithmic}[1]
\Require training dataset $\mathcal{D}_{\text{train}}$, generated dataset $\mathcal{D}_{\text{synth}}$, holdout set $\mathcal{D}_{\text{holdout}}$, number of samples in auxiliary dataset $N_\text{aux}$, privacy attack $\mathcal{A}$
\Ensure $ \mathcal{D}_{\text{train}},\ \mathcal{D}_{\text{synth}}, \  \mathcal{D}_{\text{holdout}}$ are pairwise disjoint
\State Sample $\mathcal{D}_{\text{aux}} = \left\{\mathbf{d}_i : \mathbf{d}_i \in \mathcal{D}_{\text{holdout}},  i \leq N_\text{aux} \right\}$
\State Sample $\mathcal{D}_{\text{test}} = \left\{\mathbf{d}_i : \mathbf{d}_i \in \mathcal{D}_{\text{holdout}}, \ \mathbf{d}_i \notin \mathcal{D}_{\text{aux}},\ N_\text{aux} < i \leq N_\text{train} + N_\text{aux} \right\} \cup \mathcal{D}_{\text{train}}$
\State Train $\mathcal{A}$ on $\mathcal{D}_{\text{synth}}$ and/or $\mathcal{D}_{\text{aux}}$ \label{alg:train-synth}
\State Compute scores  $\mathcal{A}(\mathbf{d}_i)$ for all $\mathbf{d}_i \in \mathcal{D}_{\text{test}}$
\State Compute threshold $\tau = \operatorname{median}(\left\{\mathcal{A}(\mathbf{d}_i) : \mathbf{d}_i \in \mathcal{D}_{\text{test}} \right\})$
\State Infer labels $m_i = \begin{cases} 1 & \text{if } \mathcal{A}(\mathbf{d}_i) > \tau \\ 0 & \text{if } \mathcal{A}(\mathbf{d}_i) \leq \tau \end{cases}, \quad \forall \mathbf{d}_i \in \mathcal{D}_{\text{test}}$
\State \Return Predicted labels and scores $M = \{(m_i, \mathcal{A}(\mathbf{d}_i)) : \forall \mathbf{d}_i \in \mathcal{D}_{\text{test}}, i < |\mathcal{D}_{\text{test}}|\}$
\end{algorithmic}
\end{algorithm}
To evaluate the utility of auxiliary data that are distinct from the source yet statistically similar, we conducted experiments using an external dataset. We selected the eICU database because the US hospitals contributing to eICU and MIMIC-IV operate under comparable protocols and utilize similar ICU management software. Consequently, $\mathcal{D}^\text{eICU}$ served as the auxiliary dataset $\mathcal{D}_\text{aux}$ for these experiments.

Finally, to establish a baseline performance for membership inference attacks, we conceived two scenarios as sanity checks, where training data from synthetic data generators were used as a direct input to the membership inference attacks. So, $\mathcal{D}_\text{train}$ was used as $\mathcal{D}_\text{synth}$. In the first scenario, 100\% of $\mathcal{D}_\text{train}$ was used to train an attack, see ~\Cref{alg:train-synth} in Algortihm~\Cref{alg:workflow}. In the second scenario, only 80\% of $\mathcal{D}_\text{train}$ was used to train an attack. As with previous experiments, sanity checks were conducted in two distinct settings: timeseries only and patient attributes only.

\paragraph{Dimensionality reduction}
The time-series datasets in the study were processed to include 9 features and 288 time points with a varying number of records as discussed in ~\Cref{sec:data-preparation}.
Given the high-dimensional nature of our dataset compared to the number of available records, it is beneficial to reduce the number of dimensions for subsequent ML modeling.
Following \citet{hilprechtMonteCarloReconstruction2019}, principal component analysis~\citep[PCA,][]{tipping_mixtures_1999} was used to decompose the data into orthogonal components that explain the maximum amount of variance. Features and time indices were treated as separate dimensions, so that the final data matrix submitted to PCA had the form samples $\times$ (features $*$ time points). Default settings of the scikit-learn library\citep{scikit-learn} were used, so that 40 PCA components were retained as in \citep{hilprechtMonteCarloReconstruction2019}, explaining 0.99\% of variance. PCA projections were estimated on the synthetic data and then applied to the other datasets used in the attack. PCA was preapplied in the following attacks when attacking time series: $\mathcal{A}_{MC-\theta}$ (\Cref{eq:mc-attack}), both variants of $\mathcal{A}_{\text{GAN-Leak}}$ (\Cref{eq:gan-leak-breugel} and \Cref{eq:gan-leak}), $\mathcal{A}_{\text{GAN-Leak-cal}}$ (\Cref{eq:gan-leak-cal}), $\mathcal{A}_{\text{DOMIAS-eq1}}$ (\Cref{eq:domias-eq1}), and $\mathcal{A}_{\text{DOMIAS-eq2}}$ (\Cref{eq:domias-eq2}) with Gaussian kernel density estimation.

\paragraph{MIA performance metrics}
\label{sec:metrics}
The performance of the attacks was evaluated using accuracy, true positive rate (TPR), false positive rate (FPR), and the area under the receiver operating characteristic curve (AUROC). Accuracy is the fraction of correctly classified samples:
\begin{align}
    \label{eq:accuracy}
    \text{Accuracy} & = \frac{1}{n}\sum_n {\mathbf{1}_{m_i^{true} = m_i^{pred}}}
\end{align}
where $\mathbf{1}_{m_i^{true} = m_i^{pred}}$ is an indicator function, equal to 1 when the predicted label of record $i$ is equal to the true label.
The area under the ROC curve (AUROC) represents the likelihood that the score assigned by the attacker to the random sample that was used in training of the generative model is higher compared to a sample that was not used:
\begin{align}
    \label{eq:auroc-u}
    \text{AUROC}(\mathcal{A}, \mathcal{D}_\text{not-train}\cup\mathcal{D}_\text{train}) & = \frac{\sum_{\mathbf{d}_i \in \mathcal{D}_\text{not-train}}\sum_{\mathbf{d}_j \in \mathcal{D}_\text{train}}\mathbf{1}[\mathcal{A}(\mathbf{d}_i) < \mathcal{A}(\mathbf{d}_j)]}{|\mathcal{D}_\text{not-train}| * |\mathcal{D}_\text{train}|} \;,
\end{align}

where $\mathcal{D}_\text{not-train} = \mathcal{D}_\text{test}\setminus \mathcal{D}_\text{train}$ is a part of $\mathcal{D}_\text{test}$ that was not used in training of synthetic data generator, $\mathbf{1}[\mathcal{A}(\mathbf{d}_i) < \mathcal{A}(\mathbf{d}_j)]$ is an indicator function, equal to 1 when if $\mathcal{A}(\mathbf{d}_i) < \mathcal{A}(\mathbf{d}_j)$, and otherwise it is 0.
The performance of a random classifier is characterized by AUROC $= 0.5$.

To measure the degree to which synthetic data overfit to the training data, we also evaluated the normalized root mean square
\begin{align}
\operatorname{NRMSE_{\operatorname{min}}}(\mathcal{D}_\text{synth}) = \operatorname{NRMSE_{\operatorname{min}}(\mathcal{D}_{\text{holdout}}, \mathcal{D}_{\text{synth}}) - NRMSE_{\operatorname{min}}(\mathcal{D}_{\text{train}}, \mathcal{D}_{\text{synth}})} \;,
\end{align}
where $\mathcal{D}_{\text{holdout}}$ is the holdout data set used for conducting the privacy attacks, $\mathcal{D}_{\text{train}}$ are the data used for training the generative model, and $\mathcal{D}_{\text{synth}}$ is a set of synthetic data generated by the model. This metric subtracts the distance between synthetic data and data from the holdout set, as well as the distance between synthetic data and data from the training set. If the generative model is overfitted, $\operatorname{NRMSE_{\operatorname{min}}}(\mathcal{D}_\text{synth})$ will be non-zero. The NRMSE between two datasets is defined as the sum over the normalized root mean square errors of pairs of samples $d_{1,t,f} \in \mathcal{D}_1$ and $d_{2,t,f} \in \mathcal{D}_2$ matched by minimal distance 
\begin{align}
\label{eq:nrmse_min}
    \operatorname{NRMSE}_{\operatorname{min}} (\mathcal{D}_{1}, \mathcal{D}_{2}) = \frac{1}{|\mathcal{D}_{1}|} \sum_{n \leq |\mathcal{D}_{1}|} \underset{j \leq |\mathcal{D}_{2}|}{\operatorname{min}} \operatorname{NRMSE}(\mathbf{d}_{n,t,f}, \mathbf{d}_{j,t,f}, \operatorname{norm}({\mathcal{D}_1, \mathcal{D}_2}))
\end{align}
where the normalization factor $\operatorname{norm}(\mathcal{D}_1, \mathcal{D}_2)$ is a $|f|$-tuple of differences between max and min values of clinical variables of two datasets, $\mathcal{D}_1$ and $\mathcal{D}_2$, defined as

\begin{align}
    \operatorname{norm}({\mathcal{D}_1, \mathcal{D}_2}) = \operatorname{max}_f(\mathcal{D}_1 \cup \mathcal{D}_2) - \operatorname{min}_f(\mathcal{D}_1 \cup \mathcal{D}_2) \;,
\end{align}

where $\operatorname{max}_f$ and $\operatorname{min}_f$ are the maxima over the samples $n \in N$ and time indices $t_i \in T $ over both datasets for feature $f \in F$, and $\operatorname{NRMSE}(\mathbf{d}_{1,t,f}, \mathbf{d}_{2,t,f}, \operatorname{norm})$ is a normalized root mean square error between two time series $\mathbf{d}_{1,t,f}$ and $\mathbf{d}_{2,t,f}$,
\begin{align}
    \operatorname{NRMSE}(\mathbf{d}_{1,t,f}, \mathbf{d}_{2,t,f}, norm) = \frac{1}{F} \sum_{f \leq F}\frac{\sqrt{\frac{1}{T}\sum_{t \leq T} (d_{1, t, f} - d_{2, t, f})^2 }}{\operatorname{norm}} \;.
\end{align}

\subsection{Code availability}
The privacy attacks used in the experiments will be published as a separate Python package, as part of the accepted publication, and can be used as part of the model training process to assess privacy risks. The code for replication of experiments will be provided upon publication.

\section{Results} \label{sec-results}
All experiments reported here were performed on a workstation with 128 cores, 256 GB of memory, and one nVidia A100 (40GB memory) graphics processing unit. Execution of the complete experiments took approximately 7 days.

\begin{figure}[ht!]
    \centering
    \includegraphics[width=\linewidth]{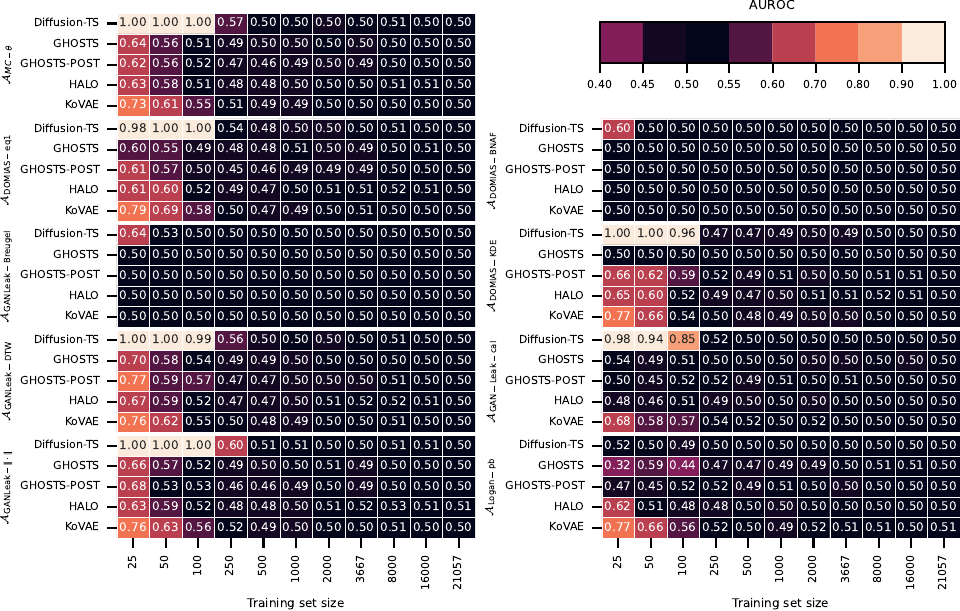}
    \caption{Performance of membership inference attacks against synthetic hospital time series generated by Diffusion-TS, GHOSTS without postprocessing (GHOSTS), GHOSTS with postprocessing (GHOSTS-POST), HALO, and KoVAE. Training set size refers to the training size of the synthetic data generator. Axes labels $\mathcal{A}_{MC-\theta}$, $\mathcal{A}_{\mathrm{DOMIAS-eq1}}$, $\mathcal{A}_{\mathrm{GAN-Leak-Breugel}}$, $\mathcal{A}_{\mathrm{GAN-Leak-DTW}}$, $\mathcal{A}_{\mathrm{GAN-Leak-\|\cdot\|}}$, $\mathcal{A}_{\mathrm{GAN-Leak-cal}}$, $\mathcal{A}_{\mathrm{DOMIAS-KDE}}$, $\mathcal{A}_{\mathrm{DOMIAS-BNAF}}$, $\mathcal{A}_{\mathrm{Logan-pb}}$ denote different privacy attacks. The left column refers to attacks that only use synthetic data, while the right column refers to attacks that use auxiliary (non-training real) data. Color-coded heat maps depict the mean of the area under the ROC curve (AUROC) estimated using $K=100$ bootstrap samples. The chance level for all attacks is AUROC = 0.5 . }
    \label{fig:timeseries}
\end{figure}

Membership inference attacks were conducted in two settings: time series only and static patient attributes only.
For the first setting, ~\Cref{fig:timeseries} summarizes the performance of all attacks on all synthetic data generators across varied training set sizes. The figure presents the mean of the AUROC metric computed using the bootstrap method (K=100). Standard errors and other metrics are available in
~\Cref{tab:tapas_df}. 
Independent of whether they rely on auxiliary (non-training real) data or not, all attacks fail to identify membership above the chance level when the training size of the synthetic data generators is above 500 samples. Below that threshold, the performance of privacy attacks diverges. For example, $\mathcal{A}_{\mathrm{GAN-Leak-Breugel}}$ and $\mathcal{A}_{\mathrm{DOMIAS-BNAF}}$ break down for generator training sample sizes around $N_\text{train} = 100$ where other attacks are still effective. The figure also shows that Diffusion-TS is more susceptible to privacy attacks in very low sample size regimes than all other generators. 

\begin{figure}
    \centering
    \includegraphics[width=\linewidth]{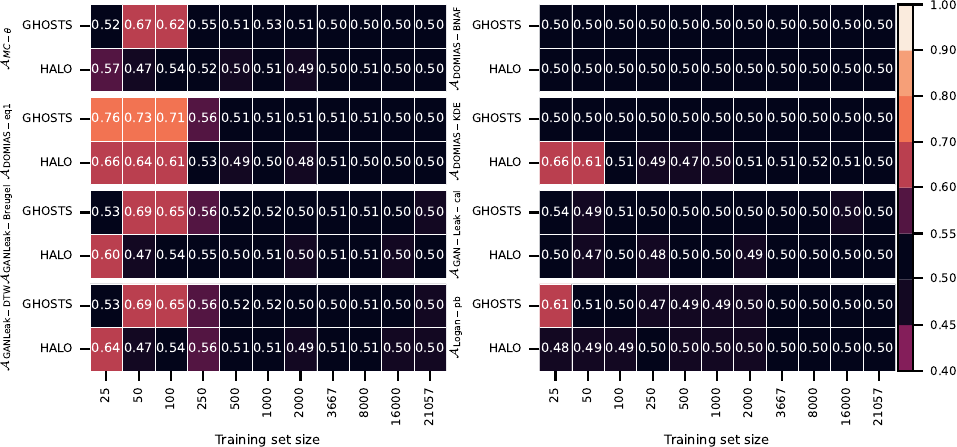}
    \caption{Performance of membership inference attacks against synthetic hospital attributes generated by GHOSTS without postprocessing (GHOSTS) and HALO. Training set size refers to the training size of the synthetic data generators. Axes labels $\mathcal{A}_{MC-\theta}$, $\mathcal{A}_{\mathrm{DOMIAS-eq1}}$, $\mathcal{A}_{\mathrm{GAN-Leak-Breugel}}$, $\mathcal{A}_{\mathrm{GAN-Leak-\|\cdot\|}}$, $\mathcal{A}_{\mathrm{GAN-Leak-cal}}$, $\mathcal{A}_{\mathrm{DOMIAS-KDE}}$, $\mathcal{A}_{\mathrm{DOMIAS-BNAF}}$, $\mathcal{A}_{\mathrm{Logan-pb}}$ denote different privacy attacks. The left column refers to attacks that only use synthetic data, while the right column refers to attacks that also use auxiliary data. Color-coded heat maps depict the mean of the area under the ROC curve (AUROC) estimated using $K=100$ bootstrap samples. The chance level for all attacks is AUROC = 0.5. }
    \label{fig:attributes}
\end{figure}

For the second setting, ~\Cref{fig:attributes} summarizes the performance of all attacks on patient attributes generated by GHOSTS and HALO only, across varied training set sizes. The figure presents the mean AUROC across $K=100$ bootstrap samples. Standard errors and other metrics are available in~\Cref{tab:tapas_attr_df}. 
The results resemble those observed for attacks against time series, although the performance of different attacks relative to each other slightly differs. Here, $\mathcal{A}_{\mathrm{GAN-Leak-cal}}$ and $\mathcal{A}_{\mathrm{DOMIAS-BNAF}}$ show significantly worse performance than other attacks in low-sample regimes. 

Sanity checks conducted in both settings for time series only show that actual training data can be identified by all attacks with high probability, although training set sizes enabling these attacks differ (\Cref{fig:sanity-check-ts}). Certain attacks, such as $\mathcal{A}_{\mathrm{DOMIAS-BNAF}}$ and $\mathcal{A}_{\mathrm{Logan-pb}}$, are ineffective against time series even in this baseline scenario for sample sizes larger than 250. $\mathcal{A}_{\mathrm{GAN-Leak-cal}}$ demonstrates above-chance performance on smaller training sizes, which drops to chance level when the training size reaches 16,000 samples. In the second scenario, where only 80\% of the training samples are available for the attacks, lower performance is naturally observed. However, $\mathcal{A}_{\mathrm{DOMIAS-eq1}}$ and $\mathcal{A}_{\mathrm{DOMIAS-KDE}}$ experience a much drastic decrease in performance than others. 

Similar conclusions can be drawn when examining the performance of the attacks against patient attributes. A notable observation is that $\mathcal{A}_{\mathrm{DOMIAS-eq1}}$ and $\mathcal{A}_{\mathrm{DOMIAS-KDE}}$ seem to be less effective against attributes in general. Standard errors and other metrics are presented in Supplementary Tables~\ref{tab:sanity_df_all}, ~\ref{tab:sanity_df_frac}, ~\ref{tab:sanity_df_attr_all}, and ~\ref{tab:sanity_df_attr_frac}.

As illustrated in \Cref{fig:eicu_comparison}, the external eICU dataset proves to be an effective source of auxiliary information for membership inference attacks. The obtained results demonstrate that attacks leveraging eICU achieve performance metrics on par with those utilizing a partition of the original dataset. Synthetic data generators that were vulnerable when only the same dataset holdout set was used received similar AUROC scores when the eICU dataset was utilized. For instance, both Diffusion-TS and KoVAE demonstrated high vulnerability in $\mathcal{A}_{\mathrm{DOMIAS-KDE}}$ and $\mathcal{A}_{\mathrm{GAN-Leak-cal}}$ attacks for both MIMIC-IV and eICU holdout sets.  Detailed results are available in ~\Cref{tab:tapas2_eicu_df}.

\begin{figure}
    \centering
    \includegraphics[width=\linewidth]{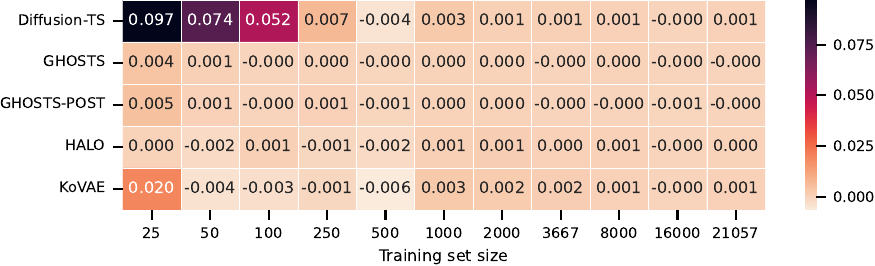}
    \caption{Estimation of overfitting using normalized root mean square metric across training sizes. Training set size refers to the training dataset size of the synthetic data generators. The values are $NRMSE_{\operatorname{min}}(\mathcal{D}_\text{syn})$, where $\mathcal{D}_\text{syn}$ is one of the datasets generated by synthetic data generators, described in \eqref{eq:nrmse_min}, estimated using bootstrapping with K=100.}
    \label{fig:ts-difference}
\end{figure}

To further investigate the privacy attacks and their connection to overfitting, we computed the mean minimal distance between the synthetic data as well as the real training and test sets, as defined in \Cref{sec:metrics} and illustrated by \Cref{fig:ts-difference}. Diffusion-TS displays the highest distance among all generative models for training set sizes below 250, with distance gradually increasing as the training set size decreases. Comparing with \Cref{fig:timeseries}, high absolute values of $NRMSE_{\operatorname{min}}(\mathcal{D}_\text{syn})$ show that the synthetic data generator is more vulnerable to MI attacks.

\section{Discussion}

Our experiments demonstrate that a wide range of membership inference attacks against synthetic hospital data are ineffective if the generators of these data are trained on more than 500 samples. This result was observed for generated time series and static attributes alike and implies that synthetically generated clinical data—including complex, high-dimensional, and multimodal time series—can be regarded as anonymous under reasonable and verifiable conditions. The training set size of 500 samples represents a moderate requirement that is attainable in most practical scenarios. Furthermore, the specific sample size requirements and associated re-identification risks can be empirically quantified through simulated privacy attacks. Consequently, this establishes a viable pathway for the secure dissemination of synthetic data for research and collaborative purposes under existing legal frameworks.

To verify our results and show a baseline of the privacy attacks, we conducted sanity checks, where attacks were trained on the actual training data of the generators. The results show that, other than $\mathcal{A}_{\mathrm{DOMIAS-BNAF}}$ and $\mathcal{A}_{\mathrm{Logan-pb}}$, all attacks are able to perform an attack with above-chance-level performance, as could be expected. Given that attacks treat synthetic data as a surrogate for the real distribution, substituting the synthetic samples with the actual training data represents the limit where surrogacy error is reduced to zero. Failing sanity check experiments may show flaws in the design of the attack.
$\mathcal{A}_{\mathrm{DOMIAS-BNAF}}$ is a variation of $\mathcal{A}_{\mathrm{DOMIAS-KDE}}$, where only the density estimation algorithm was substituted, has failed sanity check. One possible reason for its failure is, thus, the inability of BNAF to estimate the density of the time-series and static attribute data. We used implementations of DOMIAS and BNAF provided by the authors of DOMIAS, \citep{vanbreugelMembershipInferenceAttacks2023}, but we could not replicate their results on the MIMIC-IV dataset. On the other hand, $\mathcal{A}_{\mathrm{Logan-pb}}$ performs similarly to other MI attacks on some training sizes, but fails the sanity checks. Examining the performance scores, it shows that $\mathcal{A}_{\mathrm{Logan-pb}}$ maintains similarly poor scores in both \say{easy} and \say{hard} settings.

Privacy attacks against time series data are largely ineffective for the realistic attack vectors we considered when a synthetic data generator is trained on a large enough dataset. We initially attributed this phenomenon primarily to the inherent high dimensionality of time series representations. However, similar results were observed when only low-dimensional patient attributes were tested. As observed for attacks against Diffusion-TS, only those attacks that explicitly exploit distance measures between synthetic data and real data achieve performance above chance level, whereas strategies such as the $\mathcal{A}_{\mathrm{Logan-pb}}$ attack, which rely on a neural network to distinguish between auxiliary and synthetic datasets, fail to identify meaningful discriminative features.

Furthermore, as demonstrated in \Cref{fig:eicu_comparison}, utilizing the external eICU dataset as an auxiliary source for membership inference attacks yields performances comparable to that achieved using a subset of the original dataset, MIMIC-IV. Even though the two datasets represent different US hospitals, the experiments showed that one dataset can be used to break the privacy of the other. This trend suggests that the vulnerability of these generative models is not contingent on the attacker possessing samples from the exact same distribution as the training data, but rather any sufficiently representative clinical distribution. This cross-dataset efficacy highlights a critical privacy concern: an adversary does not need a "leak" from the target hospital's own database to mount a successful attack. Instead, the inherent statistical similarities between different EHR datasets (e.g., similar physiological correlations and temporal patterns) allow the attacker to construct a "reference" distribution that effectively isolates the physiological patterns that are common across hospitals. Consequently, simply restricting access to the original data partitions is insufficient to guarantee privacy if high-fidelity synthetic generators are trained on small populations, highlighting the importance of empirical privacy audit and evaluation of overfitting.

To further examine the privacy attacks and their connection to overfitting, we computed the mean minimal distance between the synthetic data and the real training and test sets. The results demonstrate that $NRMSE_{\operatorname{min}}(\mathcal{D}_\text{syn})$ is a good proxy for the success of privacy attacks. Diffusion-TS generates data that are much closer to the training samples than to samples from the holdout set, even though both data sets were drawn from the same distribution. This indicates that Diffusion-TS, trained in a low sample size regime, tends to strongly overfit, which increases the probability of success of MI attacks. As we observed similarly poor performance of MI attacks against time series and low-dimensional patient attributes alike, overfitting of the synthetic data generator appears to be a more significant factor determining the success of MIAs than data dimensionality.

Our study goes beyond prior investigations into the privacy of generative medical time series models. In the original GHOSTS publication~\citep{zhumagambetov_ghosts_2024}, the method was evaluated using $\mathcal{A}_{\mathrm{GAN-Leak-}\|\cdot\|}$. For HALO~\citep{theodorou_synthesize_2023}, privacy protection was previously assessed using MIA, attribute disclosure attacks~\citep{ye_data_free_2025} (learning private attributes from published data), and the nearest neighbour adversarial accuracy risk, a metric that estimates the degree of overfitting. In line with our results, a certain level of privacy protection was observed. The current study expands the number of MI attacks for testing and compares different time series generation methods. To our knowledge,  neither Diffusion-TS nor KoVAE have been subjected to comparable privacy audits prior to our work.

The ML community has embraced the use of differential privacy as a principled framework of privacy preservation. However, DP formulations needs to be adjusted in order to fit into the ML training process, often relaxing privacy guarantees. For example, a privacy breach found by \citet{amanyashblog2024}, described in the Introduction, analyzed synthetic data that was generated using differential privacy with claimed $\epsilon = 8$.

While various relaxations and high values of $\epsilon$ preserve the utility of synthetic data generated by DP trained models, they often fail to offer sufficient privacy protection according to the definition of differential privacy. This effectively reduces the mechanism to simple noise addition---a technique whose vulnerabilities, highlighted by \citep{dinurRevealingInformationPreserving2003}, originally motivated the development of differential privacy. Yet, the disclosure risk definitions, as discussed in the introduction, provided by DP remain valuable as they allow for the empirical validation of privacy guarantees~\citep{pmlr-v37-kairouz15, houssiauTAPASToolboxAdversarial2022}. Furthermore, there is a current consensus recommending the testing of privacy attacks against ML models. This practice not only ensures correspondence between the theoretical model and the actual implementation—preventing unaccounted leaks~\citep{blanco-justicia_critical_2022}—but also empirically validates theoretical guarantees \citep{schlegel_generating_2025}.

In this work, we performed ex-post privacy risk quantification of existing synthetic data generation algorithms using a collection of established privacy attacks, which we applied to synthetic hospital time series data. Our results demonstrate that some of the considered attacks are completely ineffective against time series, even when provided access to the actual real training data. This demonstrates a need for testing privacy attacks on realistic datasets during development. Moreover, the success of a privacy attack depends on the training dataset size; the larger it is, the better privacy protection it provides, which is contrary to DP ML practices of subsampling to amplify privacy. 

Our results suggest that generative models trained with a low number of records overfit, hence, are more susceptible to privacy attacks. A more nuanced interpretation of the results presented in \Cref{fig:ts-difference} may be that, unless synthetic data is equidistant to train and holdout datasets, it will be susceptible to privacy attacks. \citet{yeom_privacy_2018} corroborate these results, indicating that overfitting is sufficient but not necessary for privacy attacks to succeed.

This study has some limitations. First, the generative models were only trained on a single dataset, and the resulting measurement of privacy attacks' success may be different for a different choice of dataset and should not be considered universal.
Second, while multiple papers~\citep{kumar2020mlprivacy, nasr_tight_2023, annamalai_what_2024, askin_general-purpose_2025, meurers_phantom_2025, annamalai_hitchhikers_2025} have advocated for theoretical privacy guarantee validation using privacy attacks, the quantified risk only provides a lower bound of the probability of identity disclosure. Another limitation of the current study is the restricted scope of the feature space, as the evaluation was focused on time-series data or static attributes without combining longitudinal measurements with static demographic or clinical covariates. Consequently, the sample size thresholds identified in this work may shift when applied to more complex datasets. Furthermore, while dimensionality reduction was employed, the precise relationship between the number of retained PCA components and the resulting privacy-utility trade-off remains unquantified.

Existing literature generally categorizes privacy attacks against synthetic data into three primary types: membership inference, attribute disclosure, and model inversion~\citep{hittmeir_baseline_2020} (creating a copy of the synthetic data generator). In this study, we focus exclusively on Membership Inference Attacks (MIA). We excluded attribute disclosure because it relies on the prior knowledge that a target individual is present in the training set; consequently, if a generator is robust against MIA, the risk of attribute disclosure is effectively neutralized. Furthermore, attribute disclosure falls outside the standard scope of Differential Privacy (DP). Similarly, model inversion was omitted from our analysis as it reconstructs generic representative data rather than compromising the specific identities of individuals.

Notwithstanding the robust privacy implications for big enough training datasets observed in our results, empirical verification remains essential at the individual model level. We advocate for integrating analogous adversarial privacy attacks directly into the general model validation and quality assurance pipelines, as outlined in \citep{zhumagambetov_ghosts_2024}.
Moreover, further scrutiny of privacy is required for unconventional data types, such as time series. There is a need for the development and testing of privacy attacks against time series. As privacy attacks advance in complexity and effectiveness, the perceived safety margin reported here is likely to diminish in the foreseeable future. 

Furthermore, our results motivate the use of a framework of privacy attacks to estimate effective privacy rather than relying on the inherent stochastic nature of synthetic data generators or prematurely applying differential privacy where it might not add effective privacy. There is also a need to assess privacy across training set sizes.

\section{Conclusion}

To summarize, we have presented a privacy audit of synthetic time series data generators trained on public hospital data from intensive care units. As demonstrated in our experiments, generated time series are not vulnerable to existing privacy attacks when the training dataset size is reasonably large. Overall, the size of the training dataset of the generative model has a large impact on the efficacy of privacy attacks.

\section*{Acknowledgements}

This work has been performed within the ``Metrology for Artificial Intelligence in Medicine (M4AIM)'' programme funded by the German Federal Ministry for Economy and Climate Action (BMWK) in the frame of the QI-Digital Initiative.

\section*{Data Availability}
Synthetic data produced for these experiments will be submitted to Physionet.org~\citep{goldberger_physiobank_2000} for storage under MIMIC-IV's data usage agreement.

\section*{Additional Information}
Sebastian Boie reports salaried employment at Pfizer Pharma GmbH. Pfizer had no involvement in the conception, design, execution, or interpretation of the study, nor in the preparation or decision to submit the manuscript for publication. We do not report further conflicts of interest. 

\putbib

\end{bibunit}

\section*{Author contributions statement}
Z.R. and S.H. conceived the study. Z.R. developed the theoretical framework and performed the
experiments. N.G. and S.B. aided in the analysis. S.H. supervised the project. All authors discussed
the results and contributed to the final manuscript.

\begin{bibunit}
\newpage
\onecolumn

\begin{center}
{\Large Generative clinical time series models trained on moderate amounts of patient data are privacy preserving}

\vspace{0.5cm}

{Rustam Zhumagambetov, Niklas Giesa, Sebastian D. Boie, Stefan Haufe}
\end{center}

\setcounter{section}{0}
\setcounter{equation}{0}
\setcounter{figure}{0}
\setcounter{table}{0}
\setcounter{page}{1}
\renewcommand{\theequation}{S\arabic{equation}}
\renewcommand{\thefigure}{S\arabic{figure}}
\renewcommand{\thetable}{S\arabic{table}}
\renewcommand{\thesection}{S\arabic{section}}
\crefalias{figure}{appendixfigure}%
\crefalias{table}{appendixtable}%
\renewcommand{\figurename}{Supplementary Figure}
\renewcommand{\tablename}{Supplementary Table}
\makeatletter

\appendix

\section{Feature description}
\label{sec:feature-description}
\begin{table}[ht!]
        \centering
    \sisetup{%
         round-mode=places,round-precision=2,
         }
                 \caption{List of features extracted from the MIMIC-IV and eICU databases. Min and Max columns refer to the minimal and maximal values of the preprocessed data after outlier removal.}
\begin{tabular}{@{}l S[table-format=2.3] S[table-format=3.0] S[table-format=4.1] S[table-format=3.1] l@{}}
\toprule
\multirow{2}{*}{Feature name} & \multicolumn{2}{c}{Min} & \multicolumn{2}{c}{Max} & \multirow{2}{*}{Unit} \\ \cmidrule(lr){2-3} \cmidrule(lr){4-5}
                              & {MIMIC-IV}     & {eICU}     & {MIMIC-IV}     & {eICU}     &                       \\\midrule
Diastolic blood pressure      & \num{1.0    }      & \num{0  }      & \num{293.0 }       & \num{199.5}    & mmHg                   \\
Systolic blood pressure       & \num{0.202  }      & \num{0  }      & \num{365.0 }       & \num{300.0}    & mmHg                   \\
Respiratory rate              & \num{1.0    }      & \num{0  }      & \num{69.0  }       & \num{120.0}    & bpm                   \\
Heart rate                    & \num{1.0    }      & \num{0  }      & \num{295.0 }       & \num{300.0}    & bpm                   \\
SpO2                          & \num{0.300  }      & \num{50 }      & \num{100.0 }       & \num{100  }    & \%                    \\
SOFA score                    & \num{0      }      & \num{0  }      & \num{23    }       & \num{23   }    & point                 \\
Glucose                       & \num{33.000 }      & \num{33 }      & \num{1894.0}       & \num{993  }    & mg/dL                 \\
Sodium                        & \num{77.000 }      & \num{113}      & \num{186.0 }       & \num{165  }    & mmol/L                \\
Temperature                   & \num{26.100 }      & \num{32 }      & \num{42.3  }       & \num{42   }    & °C                     \\ \bottomrule
\end{tabular}
\label{tab:features}
\end{table}

\section{Data preparation}
\label{sec:data-preparation}
An ICU stay was defined as the period between the initial and last heart rate monitor recordings. We analyzed signals collected during the first 48 hours of each ICU stay. Stays with any missing ICU time series data or a total duration shorter than 48 hours were excluded, resulting in 46,337 ICU stays from 35,794 patients for the MIMIC-IV dataset, and 8,348 ICU stays from 8,163 patients for the eICU dataset.

We additionally collected static patient characteristics, such as age, marital status, length of stay, reported race and gender, along with the Sequential Organ Failure Assessment (SOFA) score~\citep{vincent_sofa_1996}, which serves as a measure of illness severity. Predicting the SOFA score constitutes the downstream task used to evaluate the usefulness of the generated synthetic data. The SOFA scores were obtained through SQL queries provided within the MIMIC IV database~\citep{johnson_mimic-iv_2023} and were computed hourly from 24 to 48 hours after ICU admission, based on data from the preceding 24-hour period.

To prepare the data for ML model training, the irregular sampling rates of individual measurements needed to be standardized. Using a sufficiently fine fixed sampling grid allowed irregularly spaced data to appear as stepwise constant time series, with steps occurring at uneven intervals. The 48-hour recordings were upsampled by assigning each 10-minute interval the value of the most recent preceding measurement and then interpolated to a uniform 10-minute sampling rate using forward filling. This process yielded $|T| = 288$ time points, where $T = {t_i}$ represents the sequence of time indices. Subsequently, we applied forward filling followed by median imputation (a median estimated from a holdout set of 1,000 patients (1,283 ICU stays)) to address missing values, resulting in a final cohort of 34,641 patients from the MIMIC-IV dataset. Median imputation with the same parameters was then applied to the eICU dataset.

\clearpage

\section{Figures}

\begin{figure}
    \centering
    \includegraphics[width=\linewidth]{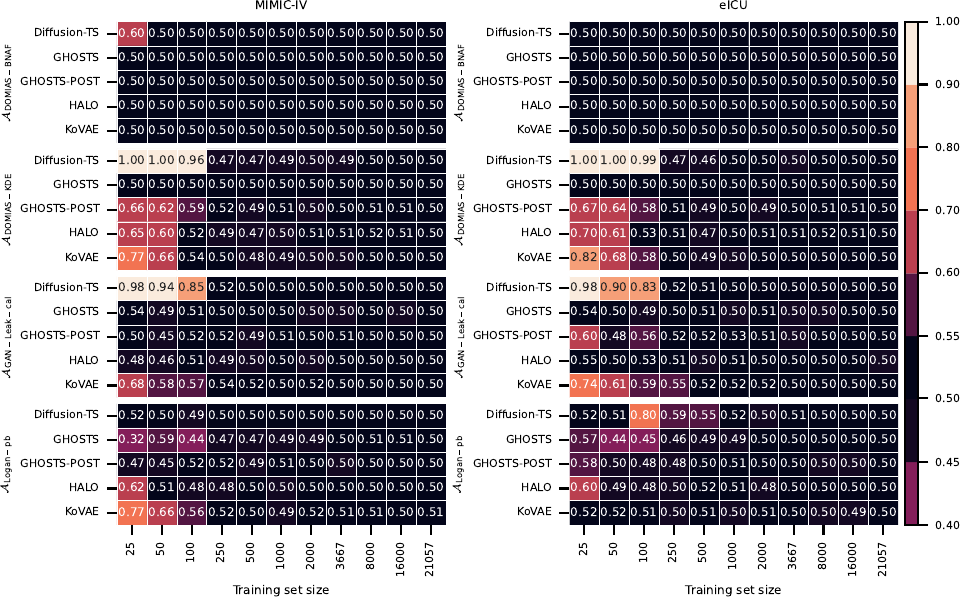}
    \caption{Performance of membership inference attacks, when eICU data was used to mount an attack, against synthetic hospital time series generated by Diffusion-TS, GHOSTS without postprocessing (GHOSTS), GHOSTS with postprocessing (GHOSTS-POST), HALO, and KoVAE. Training set size refers to the training size of the synthetic data generator. Axes labels $\mathcal{A}_{\mathrm{GAN-Leak-cal}}$, $\mathcal{A}_{\mathrm{DOMIAS-KDE}}$, $\mathcal{A}_{\mathrm{DOMIAS-BNAF}}$, $\mathcal{A}_{\mathrm{Logan-pb}}$ denote different privacy attacks. The left column refers to attacks that use part of MIMIC-IV as auxiliary information. The right column refers to attacks that use eICU as auxiliary information. Color-coded heat maps depict the mean of the area under the ROC curve (AUROC) estimated using $K=100$ bootstrap samples. The chance level for all attacks is AUROC = 0.5.}
    \label{fig:eicu_comparison}
\end{figure}

\clearpage

\section{Sanity checks}
\label{sec:sanity-check}
This section presents the results of the sanity check experiments described in \Cref{sec:privacy-assessment}.

\begin{figure}
    \centering
    \includegraphics[width=\linewidth]{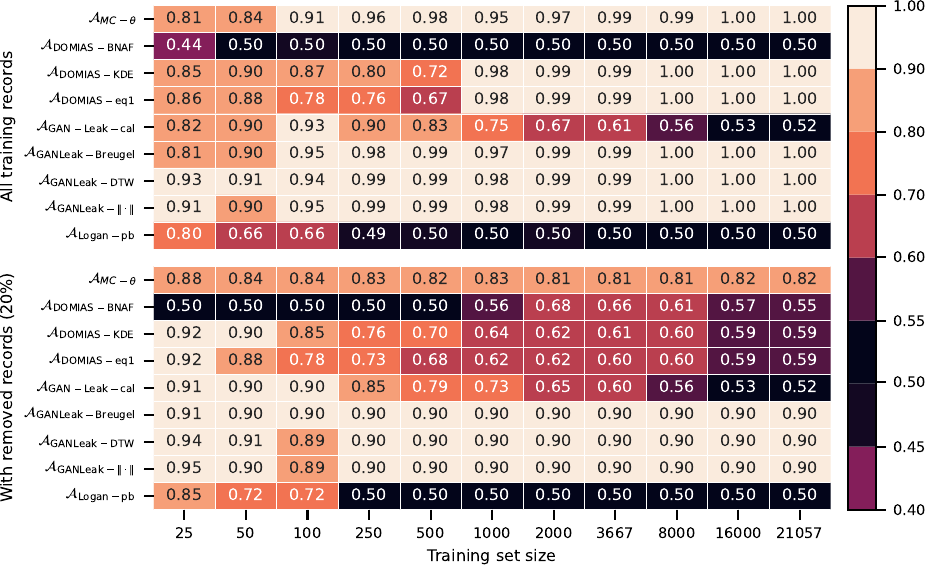}
    \caption{Performance of privacy attacks against time series ground truth data. Training set size refers to the size of real data used to train the attack. $\mathcal{A}_{MC-\theta}$, $\mathcal{A}_{\mathrm{DOMIAS-eq1}}$, $\mathcal{A}_{\mathrm{GAN-Leak-Breugel}}$, $\mathcal{A}_{\mathrm{GAN-Leak-\|\cdot\|}}$, $\mathcal{A}_{\mathrm{GAN-Leak-cal}}$, $\mathcal{A}_{\mathrm{DOMIAS-KDE}}$, $\mathcal{A}_{\mathrm{DOMIAS-BNAF}}$, $\mathcal{A}_{\mathrm{Logan-pb}}$ are the privacy attacks. The upper figure shows the scenario where 100\% of the test labeled as training size members was used for training of the attacks. The lower figure shows the scenario where only 80\% of the test labeled as training size member was used for training of the attacks. The values are the mean of the area under the ROC curve (AUROC) estimated using bootstrapping with K=100. The chance level for all attacks is AUROC = 0.5. }
    \label{fig:sanity-check-ts}
\end{figure}

\begin{figure}
    \centering
    \includegraphics[width=\linewidth]{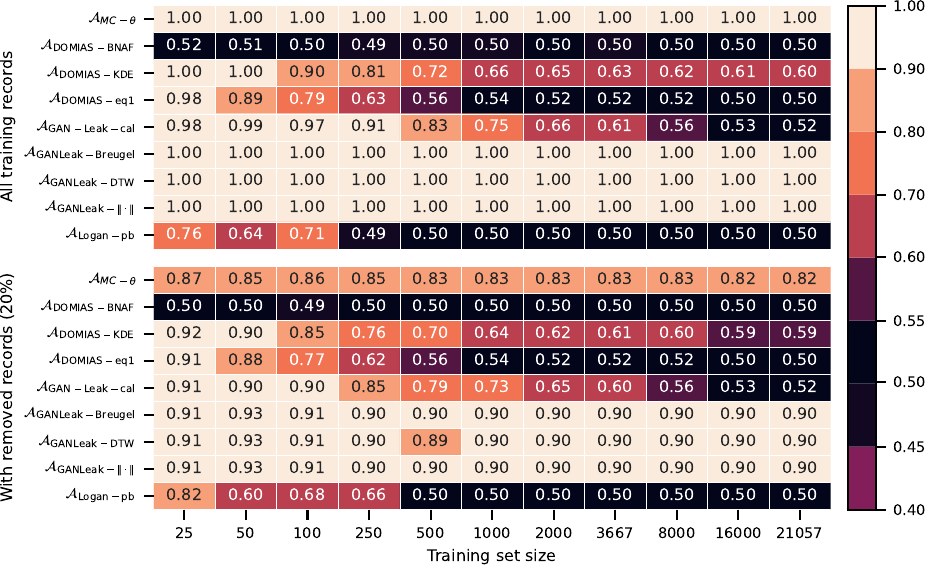}
    \caption{Performance of privacy attacks against attributes ground truth data. Training set size refers to the size of real data used to train the attack. $\mathcal{A}_{MC-\theta}$, $\mathcal{A}_{\mathrm{DOMIAS-eq1}}$, $\mathcal{A}_{\mathrm{GAN-Leak-Breugel}}$, $\mathcal{A}_{\mathrm{GAN-Leak-\|\cdot\|}}$, $\mathcal{A}_{\mathrm{GAN-Leak-cal}}$, $\mathcal{A}_{\mathrm{DOMIAS-KDE}}$, $\mathcal{A}_{\mathrm{DOMIAS-BNAF}}$, $\mathcal{A}_{\mathrm{Logan-pb}}$ are the privacy attacks. The upper figure shows the scenario where 100\% of the test labeled as training size members was used for training of the attacks. The lower figure shows the scenario where only 80\% of the test labeled as training size member was used for training of the attacks. The values are the mean of the area under the ROC curve (AUROC) estimated using bootstrapping with K=100. The chance level for all attacks is AUROC = 0.5. }
    \label{fig:sanity-check-attr}
\end{figure}

\newpage

\newgeometry{left=1.5cm,bottom=0.1cm}
\section{Tables}

\begin{longtable}{llllll}
\caption{Quantitative evaluation of how effective privacy attacks are against ground truth time series data. Train size refers to the size of real data used to train the attack. $\mathcal{A}_{MC-\theta}$, $\mathcal{A}_{\mathrm{DOMIAS-eq1}}$, $\mathcal{A}_{\mathrm{GAN-Leak-Breugel}}$, $\mathcal{A}_{\mathrm{GAN-Leak-\|\cdot\|}}$, $\mathcal{A}_{\mathrm{GAN-Leak-DTW}}$, $\mathcal{A}_{\mathrm{GAN-Leak-cal}}$, $\mathcal{A}_{\mathrm{DOMIAS-KDE}}$, $\mathcal{A}_{\mathrm{DOMIAS-BNAF}}$, $\mathcal{A}_{\mathrm{Logan-pb}}$ are the privacy attacks. The table shows the scenario where 100\% of the test test labeled as training size member was used for training of the attacks. The values of the mean and standard deviation are estimated using bootstrapping with K=100.} \label{tab:sanity_df_all} \\
\toprule
{} & {}  & {Accuracy} & {TPR} & {FPR} & {AUROC} \\
{Attack} & {Train size} & {} & {} & {} & {} \\
\midrule
\endfirsthead

\toprule
{} & {}  & {Accuracy} & {TPR} & {FPR} & {AUROC} \\
{Attack} & {Train size} & {} & {} & {} & {} \\
\midrule
\endhead
\midrule
\multicolumn{6}{r}{Continued on next page} \\
\midrule
\endfoot
\bottomrule
\endlastfoot
\multirow[c]{10}{*}{$\mathcal{A}_{MC-\theta}$} & 25 & \num{0.815} $\pm$ \num{0.063} & \num{0.809} $\pm$ \num{0.064} & \num{0.178} $\pm$ \num{0.063} & \num{0.813} $\pm$ \num{0.066} \\
\cline{2-6}
 & 50 & \num{0.846} $\pm$ \num{0.042} & \num{0.841} $\pm$ \num{0.043} & \num{0.149} $\pm$ \num{0.043} & \num{0.843} $\pm$ \num{0.05} \\
\cline{2-6}
 & 100 & \num{0.909} $\pm$ \num{0.027} & \num{0.907} $\pm$ \num{0.027} & \num{0.089} $\pm$ \num{0.028} & \num{0.907} $\pm$ \num{0.029} \\
\cline{2-6}
 & 250 & \num{0.964} $\pm$ \num{0.012} & \num{0.963} $\pm$ \num{0.012} & \num{0.035} $\pm$ \num{0.012} & \num{0.963} $\pm$ \num{0.012} \\
\cline{2-6}
 & 500 & \num{0.98} $\pm$ \num{0.007} & \num{0.98} $\pm$ \num{0.007} & \num{0.019} $\pm$ \num{0.007} & \num{0.98} $\pm$ \num{0.007} \\
\cline{2-6}
 & 1000 & \num{0.953} $\pm$ \num{0.007} & \num{0.953} $\pm$ \num{0.007} & \num{0.047} $\pm$ \num{0.007} & \num{0.948} $\pm$ \num{0.008} \\
\cline{2-6}
 & 2000 & \num{0.975} $\pm$ \num{0.003} & \num{0.975} $\pm$ \num{0.003} & \num{0.025} $\pm$ \num{0.003} & \num{0.973} $\pm$ \num{0.004} \\
\cline{2-6}
 & 3667 & \num{0.986} $\pm$ \num{0.002} & \num{0.986} $\pm$ \num{0.002} & \num{0.014} $\pm$ \num{0.002} & \num{0.985} $\pm$ \num{0.002} \\
\end{longtable}
 \newpage
\begin{longtable}{llllll}
\caption{Quantitative evaluation of how effective privacy attacks are against ground truth time series data. Train size refers to the size of real data used to train the attack. $\mathcal{A}_{MC-\theta}$, $\mathcal{A}_{\mathrm{DOMIAS-eq1}}$, $\mathcal{A}_{\mathrm{GAN-Leak-Breugel}}$, $\mathcal{A}_{\mathrm{GAN-Leak-\|\cdot\|}}$, $\mathcal{A}_{\mathrm{GAN-Leak-DTW}}$, $\mathcal{A}_{\mathrm{GAN-Leak-cal}}$, $\mathcal{A}_{\mathrm{DOMIAS-KDE}}$, $\mathcal{A}_{\mathrm{DOMIAS-BNAF}}$, $\mathcal{A}_{\mathrm{Logan-pb}}$ are the privacy attacks. The table shows the scenario where 80\% of the test test labeled as training size member was used for training of the attacks. The values of the mean and standard deviation are estimated using bootstrapping with K=100.} \label{tab:sanity_df_frac} \\
\toprule
{} & {} & {Accuracy} & {TPR} & {FPR} & {AUROC} \\
{Attack} & {Train size} & {} & {} & {} & {} \\
\midrule
\endfirsthead

\toprule
{} & {} & {Accuracy} & {TPR} & {FPR} & {AUROC} \\
{Attack} & {Train size} & {} & {} & {} & {} \\
\midrule
\endhead
\midrule
\multicolumn{6}{r}{Continued on next page} \\
\midrule
\endfoot
\bottomrule
\endlastfoot
\multirow[c]{10}{*}{$\mathcal{A}_{MC-\theta}$} & 25 & \num{0.88} $\pm$ \num{0.055} & \num{0.872} $\pm$ \num{0.056} & \num{0.113} $\pm$ \num{0.057} & \num{0.876} $\pm$ \num{0.057} \\
\cline{2-6}
 & 50 & \num{0.846} $\pm$ \num{0.042} & \num{0.841} $\pm$ \num{0.043} & \num{0.149} $\pm$ \num{0.043} & \num{0.843} $\pm$ \num{0.05} \\
\cline{2-6}
 & 100 & \num{0.844} $\pm$ \num{0.029} & \num{0.842} $\pm$ \num{0.03} & \num{0.154} $\pm$ \num{0.028} & \num{0.839} $\pm$ \num{0.032} \\
\cline{2-6}
 & 250 & \num{0.84} $\pm$ \num{0.017} & \num{0.839} $\pm$ \num{0.018} & \num{0.159} $\pm$ \num{0.017} & \num{0.833} $\pm$ \num{0.02} \\
\cline{2-6}
 & 500 & \num{0.834} $\pm$ \num{0.015} & \num{0.833} $\pm$ \num{0.015} & \num{0.165} $\pm$ \num{0.015} & \num{0.825} $\pm$ \num{0.017} \\
\cline{2-6}
 & 1000 & \num{0.837} $\pm$ \num{0.009} & \num{0.837} $\pm$ \num{0.009} & \num{0.162} $\pm$ \num{0.009} & \num{0.825} $\pm$ \num{0.009} \\
\cline{2-6}
 & 2000 & \num{0.832} $\pm$ \num{0.006} & \num{0.832} $\pm$ \num{0.006} & \num{0.168} $\pm$ \num{0.006} & \num{0.812} $\pm$ \num{0.007} \\
\cline{2-6}
 & 3667 & \num{0.828} $\pm$ \num{0.005} & \num{0.828} $\pm$ \num{0.005} & \num{0.172} $\pm$ \num{0.005} & \num{0.81} $\pm$ \num{0.006} \\
\end{longtable}
 \newpage
\begin{longtable}{llllll}
\caption{Quantitative evaluation of how effective privacy attacks are against ground truth attributes data. Train size refers to the size of real data used to train the attack. $\mathcal{A}_{MC-\theta}$, $\mathcal{A}_{\mathrm{DOMIAS-eq1}}$, $\mathcal{A}_{\mathrm{GAN-Leak-Breugel}}$, $\mathcal{A}_{\mathrm{GAN-Leak-\|\cdot\|}}$, $\mathcal{A}_{\mathrm{GAN-Leak-cal}}$, $\mathcal{A}_{\mathrm{DOMIAS-KDE}}$, $\mathcal{A}_{\mathrm{DOMIAS-BNAF}}$, $\mathcal{A}_{\mathrm{Logan-pb}}$ are the privacy attacks. The table shows the scenario where 100\% of the test test labeled as training size member was used for training of the attacks. The values of the mean and standard deviation are estimated using bootstrapping with K=100.} \label{tab:sanity_df_attr_all} \\
\toprule
{} & {} & {Accuracy} & {TPR} & {FPR} & {AUROC} \\
{Attack} & {Train size} & {} & {} & {} & {} \\
\midrule
\endfirsthead

\toprule
{} & {} & {Accuracy} & {TPR} & {FPR} & {AUROC} \\
{Attack} & {Train size} & {} & {} & {} & {} \\
\midrule
\endhead
\midrule
\multicolumn{6}{r}{Continued on next page} \\
\midrule
\endfoot
\bottomrule
\endlastfoot
\multirow[c]{10}{*}{$\mathcal{A}_{MC-\theta}$} & 25 & \num{1.0} $\pm$ \num{0.0} & \num{1.0} $\pm$ \num{0.0} & \num{0.0} $\pm$ \num{0.0} & \num{1.0} $\pm$ \num{0.0} \\
\cline{2-6}
 & 50 & \num{1.0} $\pm$ \num{0.0} & \num{1.0} $\pm$ \num{0.0} & \num{0.0} $\pm$ \num{0.0} & \num{1.0} $\pm$ \num{0.0} \\
\cline{2-6}
 & 100 & \num{1.0} $\pm$ \num{0.0} & \num{1.0} $\pm$ \num{0.0} & \num{0.0} $\pm$ \num{0.0} & \num{1.0} $\pm$ \num{0.0} \\
\cline{2-6}
 & 250 & \num{1.0} $\pm$ \num{0.0} & \num{1.0} $\pm$ \num{0.0} & \num{0.0} $\pm$ \num{0.0} & \num{1.0} $\pm$ \num{0.0} \\
\cline{2-6}
 & 500 & \num{1.0} $\pm$ \num{0.0} & \num{1.0} $\pm$ \num{0.0} & \num{0.0} $\pm$ \num{0.0} & \num{1.0} $\pm$ \num{0.0} \\
\cline{2-6}
 & 1000 & \num{1.0} $\pm$ \num{0.0} & \num{1.0} $\pm$ \num{0.0} & \num{0.0} $\pm$ \num{0.0} & \num{1.0} $\pm$ \num{0.0} \\
\cline{2-6}
 & 2000 & \num{1.0} $\pm$ \num{0.0} & \num{1.0} $\pm$ \num{0.0} & \num{0.0} $\pm$ \num{0.0} & \num{1.0} $\pm$ \num{0.0} \\
\cline{2-6}
 & 3667 & \num{1.0} $\pm$ \num{0.0} & \num{1.0} $\pm$ \num{0.0} & \num{0.0} $\pm$ \num{0.0} & \num{1.0} $\pm$ \num{0.0} \\
\end{longtable}
 \newpage
\begin{longtable}{llllll}
\caption{Quantitative evaluation of how effective privacy attacks are against ground truth attributes data. Train size refers to the size of real data used to train the attack. $\mathcal{A}_{MC-\theta}$, $\mathcal{A}_{\mathrm{DOMIAS-eq1}}$, $\mathcal{A}_{\mathrm{GAN-Leak-Breugel}}$, $\mathcal{A}_{\mathrm{GAN-Leak-\|\cdot\|}}$, $\mathcal{A}_{\mathrm{GAN-Leak-cal}}$, $\mathcal{A}_{\mathrm{DOMIAS-KDE}}$, $\mathcal{A}_{\mathrm{DOMIAS-BNAF}}$, $\mathcal{A}_{\mathrm{Logan-pb}}$ are the privacy attacks. The table shows the scenario where 80\% of the test test labeled as training size member was used for training of the attacks. The values of the mean and standard deviation are estimated using bootstrapping with K=100.} \label{tab:sanity_df_attr_frac} \\
\toprule
{} & {} & {Accuracy} & {TPR} & {FPR} & {AUROC} \\
{Attack} & {Train size}& {} & {} & {} & {} \\
\midrule
\endfirsthead

\toprule
{} & {} & {Accuracy} & {TPR} & {FPR} & {AUROC} \\
{Attack} & {Train size} & {} & {} & {} & {} \\
\midrule
\endhead
\midrule
\multicolumn{6}{r}{Continued on next page} \\
\midrule
\endfoot
\bottomrule
\endlastfoot
\multirow[c]{10}{*}{$\mathcal{A}_{MC-\theta}$} & 25 & \num{0.83} $\pm$ \num{0.069} & \num{0.82} $\pm$ \num{0.069} & \num{0.16} $\pm$ \num{0.072} & \num{0.872} $\pm$ \num{0.054} \\
\cline{2-6}
 & 50 & \num{0.844} $\pm$ \num{0.042} & \num{0.838} $\pm$ \num{0.044} & \num{0.15} $\pm$ \num{0.043} & \num{0.855} $\pm$ \num{0.04} \\
\cline{2-6}
 & 100 & \num{0.86} $\pm$ \num{0.028} & \num{0.857} $\pm$ \num{0.029} & \num{0.137} $\pm$ \num{0.028} & \num{0.865} $\pm$ \num{0.028} \\
\cline{2-6}
 & 250 & \num{0.84} $\pm$ \num{0.019} & \num{0.839} $\pm$ \num{0.019} & \num{0.159} $\pm$ \num{0.019} & \num{0.846} $\pm$ \num{0.02} \\
\cline{2-6}
 & 500 & \num{0.834} $\pm$ \num{0.015} & \num{0.833} $\pm$ \num{0.015} & \num{0.166} $\pm$ \num{0.015} & \num{0.835} $\pm$ \num{0.016} \\
\cline{2-6}
 & 1000 & \num{0.836} $\pm$ \num{0.009} & \num{0.836} $\pm$ \num{0.009} & \num{0.164} $\pm$ \num{0.009} & \num{0.831} $\pm$ \num{0.01} \\
\cline{2-6}
 & 2000 & \num{0.831} $\pm$ \num{0.007} & \num{0.831} $\pm$ \num{0.007} & \num{0.169} $\pm$ \num{0.007} & \num{0.826} $\pm$ \num{0.008} \\
\cline{2-6}
 & 3667 & \num{0.833} $\pm$ \num{0.006} & \num{0.833} $\pm$ \num{0.006} & \num{0.167} $\pm$ \num{0.006} & \num{0.829} $\pm$ \num{0.006} \\
\end{longtable}
 \newpage
\begin{longtable}{lllllll}
\caption{Quantitative evaluation of how effective privacy attacks are against synthetic time series data. Train size refers to the size of real data used to train the synthetic data generator. $\mathcal{A}_{MC-\theta}$, $\mathcal{A}_{\mathrm{DOMIAS-eq1}}$, $\mathcal{A}_{\mathrm{GAN-Leak-Breugel}}$, $\mathcal{A}_{\mathrm{GAN-Leak-\|\cdot\|}}$, $\mathcal{A}_{\mathrm{GAN-Leak-DTW}}$, $\mathcal{A}_{\mathrm{GAN-Leak-cal}}$, $\mathcal{A}_{\mathrm{DOMIAS-KDE}}$, $\mathcal{A}_{\mathrm{DOMIAS-BNAF}}$, $\mathcal{A}_{\mathrm{Logan-pb}}$ are the privacy attacks. The values of the mean and standard deviation are estimated using bootstrapping with K=100.} \label{tab:tapas_df} \\
\toprule
{} & {} & {} & {Accuracy} & {TPR} & {FPR} & {AUROC} \\
{Attack} & {Train size} & {Model} & {} & {} & {} & {} \\
\midrule
\endfirsthead

\toprule
{} & {} & {} & {Accuracy} & {TPR} & {FPR} & {AUROC} \\
{Attack} & {Train size} & {Model} & {} & {} & {} & {} \\
\midrule
\endhead
\midrule
\multicolumn{7}{r}{Continued on next page} \\
\midrule
\endfoot
\bottomrule
\endlastfoot
\multirow[c]{55}{*}{$\mathcal{A}_{MC-\theta}$} & \multirow[c]{5}{*}{25} & Diffusion-TS & \num{1.0} $\pm$ \num{0.0} & \num{1.0} $\pm$ \num{0.0} & \num{0.0} $\pm$ \num{0.0} & \num{1.0} $\pm$ \num{0.0} \\
 &  & GHOSTS & \num{0.626} $\pm$ \num{0.069} & \num{0.618} $\pm$ \num{0.067} & \num{0.365} $\pm$ \num{0.073} & \num{0.639} $\pm$ \num{0.07} \\
 &  & GHOSTS-POST & \num{0.606} $\pm$ \num{0.072} & \num{0.595} $\pm$ \num{0.075} & \num{0.384} $\pm$ \num{0.073} & \num{0.625} $\pm$ \num{0.076} \\
 &  & HALO & \num{0.622} $\pm$ \num{0.077} & \num{0.612} $\pm$ \num{0.08} & \num{0.368} $\pm$ \num{0.077} & \num{0.634} $\pm$ \num{0.078} \\
 &  & KoVAE & \num{0.7} $\pm$ \num{0.07} & \num{0.689} $\pm$ \num{0.072} & \num{0.289} $\pm$ \num{0.071} & \num{0.73} $\pm$ \num{0.069} \\
\cline{2-7}
 & \multirow[c]{5}{*}{50} & Diffusion-TS & \num{1.0} $\pm$ \num{0.0} & \num{1.0} $\pm$ \num{0.0} & \num{0.0} $\pm$ \num{0.0} & \num{1.0} $\pm$ \num{0.0} \\
 &  & GHOSTS & \num{0.542} $\pm$ \num{0.051} & \num{0.538} $\pm$ \num{0.052} & \num{0.453} $\pm$ \num{0.051} & \num{0.564} $\pm$ \num{0.056} \\
 &  & GHOSTS-POST & \num{0.534} $\pm$ \num{0.051} & \num{0.528} $\pm$ \num{0.051} & \num{0.46} $\pm$ \num{0.052} & \num{0.555} $\pm$ \num{0.054} \\
\end{longtable}
 \newpage
\begin{longtable}{lllllll}
\caption{Quantitative evaluation of how effective privacy attacks are against synthetic attributes data. Train size refers to the size of real data used to train the synthetic data generator. $\mathcal{A}_{MC-\theta}$, $\mathcal{A}_{\mathrm{DOMIAS-eq1}}$, $\mathcal{A}_{\mathrm{GAN-Leak-Breugel}}$, $\mathcal{A}_{\mathrm{GAN-Leak-\|\cdot\|}}$, $\mathcal{A}_{\mathrm{GAN-Leak-cal}}$, $\mathcal{A}_{\mathrm{DOMIAS-KDE}}$, $\mathcal{A}_{\mathrm{DOMIAS-BNAF}}$, $\mathcal{A}_{\mathrm{Logan-pb}}$ are the privacy attacks. The values of the mean and standard deviation are estimated using bootstrapping with K=100.} \label{tab:tapas_attr_df} \\
\toprule
{} & {} & {} & {Accuracy} & {TPR} & {FPR} & {AUROC} \\
{Attack} & {Train size} & {Model} & {} & {} & {} & {} \\
\midrule
\endfirsthead

\toprule
{} & {} & {} & {Accuracy} & {TPR} & {FPR} & {AUROC} \\
{Attack} & {Train size} & {Model} & {} & {} & {} & {} \\
\midrule
\endhead
\midrule
\multicolumn{7}{r}{Continued on next page} \\
\midrule
\endfoot
\bottomrule
\endlastfoot
\multirow[c]{22}{*}{$\mathcal{A}_{MC-\theta}$} & \multirow[c]{2}{*}{25} & GHOSTS & \num{0.48} $\pm$ \num{0.066} & \num{0.471} $\pm$ \num{0.068} & \num{0.511} $\pm$ \num{0.068} & \num{0.519} $\pm$ \num{0.069} \\
 &  & HALO & \num{0.611} $\pm$ \num{0.077} & \num{0.603} $\pm$ \num{0.076} & \num{0.38} $\pm$ \num{0.081} & \num{0.573} $\pm$ \num{0.084} \\
\cline{2-7}
 & \multirow[c]{2}{*}{50} & GHOSTS & \num{0.627} $\pm$ \num{0.048} & \num{0.623} $\pm$ \num{0.049} & \num{0.37} $\pm$ \num{0.049} & \num{0.671} $\pm$ \num{0.048} \\
 &  & HALO & \num{0.476} $\pm$ \num{0.047} & \num{0.471} $\pm$ \num{0.049} & \num{0.518} $\pm$ \num{0.047} & \num{0.472} $\pm$ \num{0.052} \\
\cline{2-7}
 & \multirow[c]{2}{*}{100} & GHOSTS & \num{0.624} $\pm$ \num{0.037} & \num{0.622} $\pm$ \num{0.036} & \num{0.375} $\pm$ \num{0.039} & \num{0.624} $\pm$ \num{0.039} \\
 &  & HALO & \num{0.542} $\pm$ \num{0.034} & \num{0.541} $\pm$ \num{0.035} & \num{0.456} $\pm$ \num{0.034} & \num{0.543} $\pm$ \num{0.039} \\
\cline{2-7}
 & \multirow[c]{2}{*}{250} & GHOSTS & \num{0.542} $\pm$ \num{0.023} & \num{0.541} $\pm$ \num{0.023} & \num{0.457} $\pm$ \num{0.023} & \num{0.545} $\pm$ \num{0.025} \\
 &  & HALO & \num{0.525} $\pm$ \num{0.024} & \num{0.524} $\pm$ \num{0.024} & \num{0.474} $\pm$ \num{0.024} & \num{0.518} $\pm$ \num{0.026} \\
\end{longtable}
 \newpage
\begin{longtable}{lllllll}
\caption{Quantitative evaluation of how effective privacy attacks are against synthetic time series data using external dataset, eICU, as auxiliary data to mount an attack. Train size refers to the size of real data used to train the synthetic data generator. $\mathcal{A}_{MC-\theta}$, $\mathcal{A}_{\mathrm{DOMIAS-eq1}}$, $\mathcal{A}_{\mathrm{GAN-Leak-Breugel}}$, $\mathcal{A}_{\mathrm{GAN-Leak-\|\cdot\|}}$, $\mathcal{A}_{\mathrm{GAN-Leak-DTW}}$, $\mathcal{A}_{\mathrm{GAN-Leak-cal}}$, $\mathcal{A}_{\mathrm{DOMIAS-KDE}}$, $\mathcal{A}_{\mathrm{DOMIAS-BNAF}}$, $\mathcal{A}_{\mathrm{Logan-pb}}$ are the privacy attacks. The values of the mean and standard deviation are estimated using bootstrapping with K=100.} \label{tab:tapas2_eicu_df} \\
\toprule
{} & {} & {} & {Accuracy} & {TPR} & {FPR} & {AUROC} \\
{Attack} & {Train size} & {Model} & {} & {} & {} & {} \\
\midrule
\endfirsthead

\toprule
{} & {} & {} & {Accuracy} & {TPR} & {FPR} & {AUROC} \\
{Attack} & {Train size} & {Model} & {} & {} & {} & {} \\
\midrule
\endhead
\midrule
\multicolumn{7}{r}{Continued on next page} \\
\midrule
\endfoot
\bottomrule
\endlastfoot
\multirow[c]{55}{*}{$\mathcal{A}_{\mathrm{DOMIAS-BNAF}}$} & \multirow[c]{5}{*}{25} & Diffusion-TS & \num{0.5} $\pm$ \num{0.0} & \num{0.0} $\pm$ \num{0.0} & \num{0.0} $\pm$ \num{0.0} & \num{0.5} $\pm$ \num{0.0} \\
 &  & GHOSTS & \num{0.5} $\pm$ \num{0.0} & \num{0.0} $\pm$ \num{0.0} & \num{0.0} $\pm$ \num{0.0} & \num{0.5} $\pm$ \num{0.0} \\
 &  & GHOSTS-POST & \num{0.5} $\pm$ \num{0.0} & \num{0.0} $\pm$ \num{0.0} & \num{0.0} $\pm$ \num{0.0} & \num{0.5} $\pm$ \num{0.0} \\
 &  & HALO & \num{0.5} $\pm$ \num{0.0} & \num{0.0} $\pm$ \num{0.0} & \num{0.0} $\pm$ \num{0.0} & \num{0.5} $\pm$ \num{0.0} \\
 &  & KoVAE & \num{0.5} $\pm$ \num{0.0} & \num{0.0} $\pm$ \num{0.0} & \num{0.0} $\pm$ \num{0.0} & \num{0.5} $\pm$ \num{0.0} \\
\cline{2-7}
 & \multirow[c]{5}{*}{50} & Diffusion-TS & \num{0.5} $\pm$ \num{0.0} & \num{0.0} $\pm$ \num{0.0} & \num{0.0} $\pm$ \num{0.0} & \num{0.5} $\pm$ \num{0.0} \\
 &  & GHOSTS & \num{0.5} $\pm$ \num{0.0} & \num{0.0} $\pm$ \num{0.0} & \num{0.0} $\pm$ \num{0.0} & \num{0.5} $\pm$ \num{0.0} \\
 &  & GHOSTS-POST & \num{0.5} $\pm$ \num{0.0} & \num{0.0} $\pm$ \num{0.0} & \num{0.0} $\pm$ \num{0.0} & \num{0.5} $\pm$ \num{0.0} \\
\end{longtable}
 \newpage
\restoregeometry \clearpage
\putbib
\end{bibunit}

\end{document}